\documentclass[journal]{IEEEtran}

\usepackage{amsmath}
\usepackage{amsfonts}
\usepackage{amssymb}

\usepackage{amsthm}
\usepackage{pifont}
\usepackage[ruled,vlined]{algorithm2e}
\usepackage{algorithmic}
\usepackage{mathtools}
\usepackage{tabularx}
\usepackage{graphicx}
\usepackage{subfigure}
\usepackage{enumerate}
\usepackage{float}
\usepackage{url}
\usepackage{verbatim}
\usepackage{cite}
\usepackage{booktabs}
\usepackage{diagbox}
\usepackage{makecell}
\usepackage{bbding}
\usepackage{xcolor}
\usepackage{arydshln}
\usepackage{multirow}
\usepackage{hhline}
\usepackage{afterpage}
\usepackage{stfloats}
\usepackage{placeins}
\usepackage[linkcolor=black,citecolor=black,urlcolor=black,colorlinks=true]{hyperref}

\graphicspath{{figures/}}
\IEEEoverridecommandlockouts

\DeclareMathOperator*{\argmax}{arg\,max}
\DeclareMathOperator*{\argmin}{arg\,min}

\DeclareMathOperator*{\diag}{diag}
\DeclareMathOperator*{\conv}{conv}
\DeclareMathOperator*{\vol}{vol}
\DeclareMathOperator*{\area }{area }

\newcommand{\tp}{^{\mathrm{T}}}

\newcommand{\rbrac}[1]{({#1})}

\newcommand{\cbrac}[1]{\{{#1}\}}
\newcommand{\cBrac}[1]{\left\{{#1}\right\}}
\newcommand{\sbrac}[1]{[{#1}]}

\newcommand{\norm}[1]{\Vert{#1}\Vert}

\newcommand{\hidden}[1]{}
\newcommand{\rv}[1]{\textcolor{black}{#1}}

\author{Qianhao Wang$^{\dag}$, Zhepei Wang$^{\dag}$, Mingyang Wang, Jialin Ji, \\Zhichao Han, Tianyue Wu, Rui Jin, Yuman Gao, Chao Xu, and Fei Gao$^*$
    \thanks{\textbf{${\dag}$ Equal contribution.}}
    \thanks{\textbf{${*}$ Corresponding author.}}
    \thanks{All authors are with the State Key Laboratory of Industrial Control Technology, Institute of Cyber-Systems and Control, Zhejiang University, Hangzhou 310027, China and Huzhou Institute, Zhejiang University, Huzhou 313000, China.  {\tt\small \{qhwangaa, wangzhepei, fgaoaa\}@zju.edu.cn}}
}

\title{Computing Large 2D/3D Convex Regions of \\ Obstacle-Free Space  for Robotic Planning}

\title{{F}ast {I}terative {R}egion {I}nflation for Computing Large 2-D/3-D Convex Regions of Obstacle-Free Space}

\begin{document}
    \maketitle

\begin{abstract}

Convex polytopes have compact representations and exhibit convexity, which makes them suitable for abstracting obstacle-free spaces from various environments.
\rv{Existing generation methods struggle with balancing high-quality output and efficiency.}
Moreover, another crucial requirement for convex polytopes to accurately contain certain seed point sets, such as a robot or a front-end path, is proposed in various tasks, which we refer to as manageability.
\rv{ 
In this paper, we propose Fast Iterative Regional Inflation (FIRI) to generate high-quality convex polytope while ensuring efficiency and manageability simultaneously.
FIRI consists of two iteratively executed submodules: Restrictive Inflation (RsI) and Maximum Volume Inscribed Ellipsoid (MVIE) computation.}
By explicitly incorporating constraints that include the seed point set,
RsI guarantees manageability.
\rv{Meanwhile, iterative MVIE optimization ensures high-quality result through monotonic volume bound improvement.}
In terms of efficiency, we design methods tailored to the low-dimensional and multi-constrained nature of both modules, resulting in orders of magnitude improvement compared to generic solvers. 
\rv{Notably, in 2-D MVIE, we present the first linear-complexity analytical algorithm for maximum area inscribed ellipse, further enhancing the performance in 2-D cases.}
Extensive benchmarks conducted against state-of-the-art methods validate the superior performance of FIRI in terms of quality, manageability, and efficiency. 
Furthermore, various real-world applications showcase the generality and practicality of FIRI. 
The high-performance code of FIRI will be open-sourced.

\end{abstract}

\section{Introduction}
\label{sec:intro}

In robotics, a key task is to navigate without collisions, which involves frequent interactions with environments abundant in vast amounts of discrete obstacle information.
\rv{For this interaction requirement, convex polytopes provide a compact, structured geometric abstraction of feasible space, alleviating the burden of collision avoidance~\cite{Deits2015ComputingIRIS,Liu2017PlanningDF,wang2022geometrically}.
Furthermore, this abstraction facilitates large-scale space storage and enables topological analysis, including the construction of road maps~\cite{guo2022dynamic} \rv{and convex covers~\cite{werner2024approximating}}}.
Additionally, convex polytopes construct convex linear safety constraints from non-convex obstacles, which benefits problem formulation and solution. 
In fact, in several applications, this approach even transforms the problem into convex optimization, leading to the attainment of global optima~\cite{Deits2015EfficientMISOS,marcucci2023motion}.

\begin{figure}[t]
    \begin{center}
        \subfigure[\label{fig:toutu large} Comparison of the impact of the convex polytopes of different quality on trajectory generation. The red curve represents the trajectory.]
        {\includegraphics[width=1.0\columnwidth]{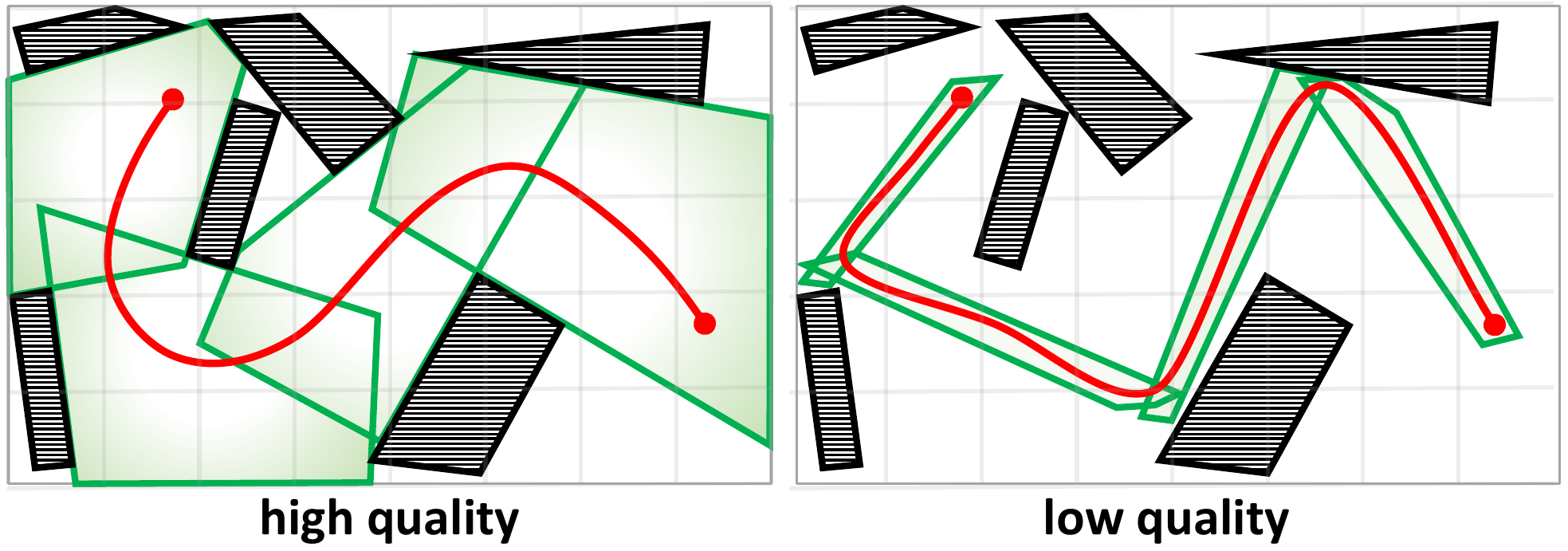}}
        \subfigure[\label{fig:toutu manageability} Comparison of polytopes with or without manageability. \textbf{Top}: When generating polytopes based on the blue lines to construct safe corridor , the absence of manageability may lead to discontinuity in the generated corridor. \textbf{Bottom}: The absence of manageability during whole-body planning may result in the failure to generate convex polytopes that contain the robot.]
        {\includegraphics[width=1.0\columnwidth]{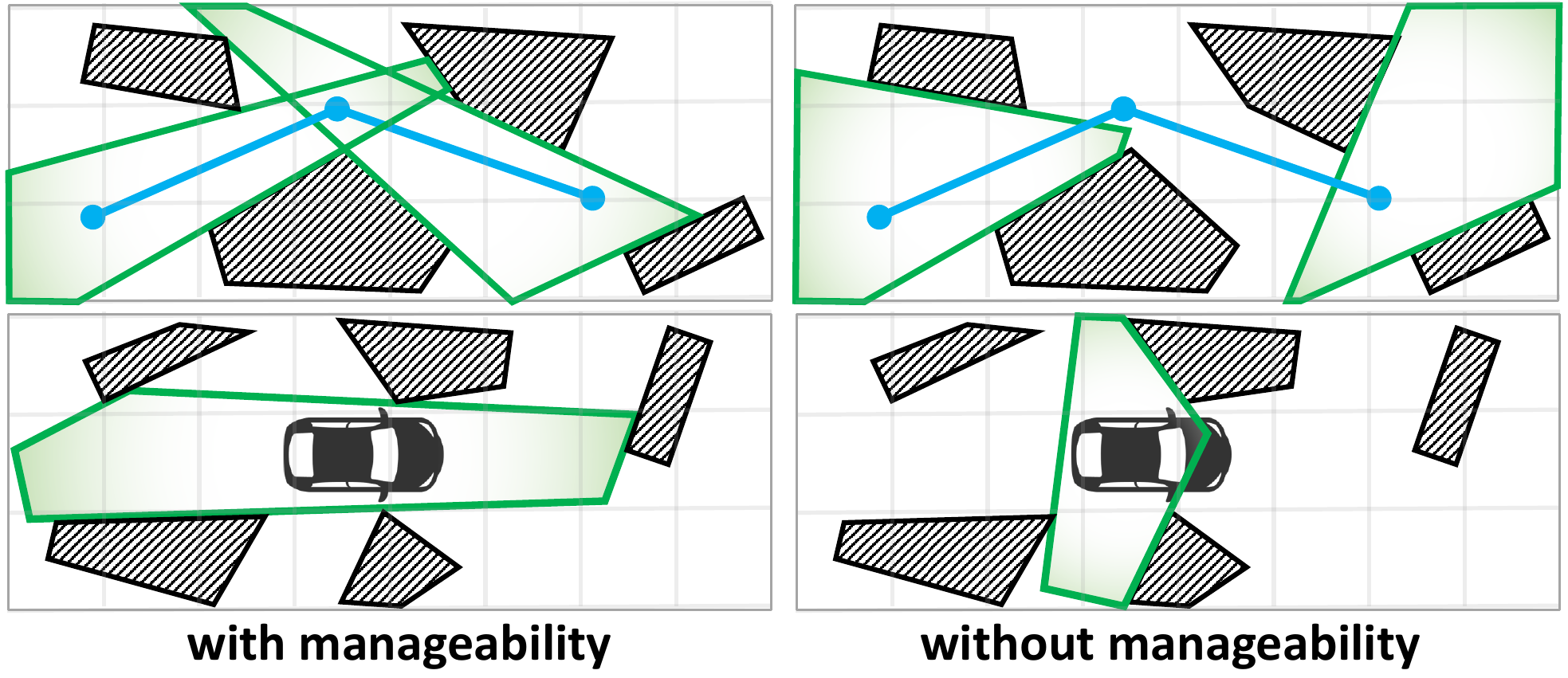}}
    \end{center}
    \vspace{-0.4cm}
    \caption{\label{fig:toutu} Illustration of quality and manageability. The gray polytopes represent obstacles, and the green polytopes represent the generated free polytopes.}
    \vspace{-0.6cm}
\end{figure}

\rv{Although convex polytopes offer compact representation, generating satisfactory ones is far from a trivial task.
First, a larger convex polytope extracts more information from the safety space, benefiting tasks that require spatial search within the safety space, such as trajectory planning.
As illustrated in Fig.~\ref{fig:toutu large}, larger convex polytopes facilitate a smoother trajectory.
Therefore, we aim to maximize the polytope volume, which we define as its \textit{quality} in this paper.
Furthermore, maximizing computational \textit{efficiency} enables applications in online high-speed tasks or with limited onboard resources.
Despite extensive research~\cite{Deits2015ComputingIRIS,Gao2020TeachRepeatReplanAC,Zhong2020GeneratingLCP,Liu2017PlanningDF}, balancing polytope quality and generation efficiency remains challenging. 
Existing approaches either generate high-quality regions but require substantial computational budget~\cite{Deits2015ComputingIRIS,Gao2020TeachRepeatReplanAC}, or achieve fast computation while yielding conservative results~\cite{Zhong2020GeneratingLCP,Liu2017PlanningDF}.}

In addition to quality and efficiency, several applications impose a crucial requirement for the generated convex polytope to accurately encompass a specified set of points, to which we refer as the \textit{manageability}. 
\rv{For example, in trajectory planning shown in Fig.~\ref{fig:toutu manageability}, the corridor-based approach~\cite{Liu2017PlanningDF,ji2022elastic} uses blue path segments as seeds to generate convex hulls, forming a safety corridor.}
If the convex hulls fail to encompass the corresponding line segments, it may result in a discontinuity in the corridor, making it impossible to generate a continue trajectory within the corridor.
\rv{Additionally, in whole-body planning~\cite{han2023efficient,wang2022geometrically},  insufficient coverage of the robot's shape can lead to similar planning failure due to the absence of feasible solution space, as illustrated in Fig.\ref{fig:toutu manageability}.}
However, most existing algorithms prioritize optimizing the size of the region without adequately considering~\cite{Liu2017PlanningDF,Deits2015ComputingIRIS} or ability to  ensure manageability~\cite{Zhong2020GeneratingLCP,savin2017algorithm}.

To satisfy these requirements, we propose a novel algorithm, called \textbf{F}ast \textbf{I}terative \textbf{R}egion \textbf{I}nflation (FIRI) for computing free convex polytope, which simultaneously achieves high \textit{quality}, high  \textit{efficiency}, and strong  \textit{manageability} for the first time.
FIRI takes obstacles, a seed consisting of points that is required to be included, and an initial ellipsoid as inputs. 
It iteratively proceeds with two modules (detailed in Fig.~\ref{fig:FIRI}): 
\begin{itemize}
    \item [1)] Restrictive Inflation (RsI): inflating the ellipsoid and using its contact planes tangent to obstacles to generate a set of halfspaces \rv{containing the seed}, which separate a convex polytope from obstacles.
    \item [2)] The Maximum Volume Inscribed Ellipsoid (MVIE) of the convex polytope is required to be calculated, which will be inflated in the next iteration.
\end{itemize}
MVIE serves as a lower bound of the volume of the convex polytope, which monotonically expands during the iterative update, leading to a growing obstacle-free region that ensures the high {quality} of FIRI.
The idea of monotonic updating the lower bound is inspired by Deits' fundamental work IRIS~\cite{Deits2015ComputingIRIS}, which, however, lacks {manageability}.
Additionally, its considerable computational overhead in complex environments~\cite{Deits2015ComputingIRIS,Zhong2020GeneratingLCP} limits its real-time applications.
\rv{In contrast, FIRI significantly improves upon IRIS in terms of manageability and computational efficiency.
For manageability, FIRI uses RsI to generate halfspaces that compose the polytope to necessarily exclude obstacles while containing the seed.
For {efficiency}, we design specialized methods for FIRI's two optimization modules based on their geometric properties, resulting in a remarkable computational efficiency improvement with a speedup of orders of magnitude compared to IRIS.}

For RsI, we convert the halfspace computation into a minimum-norm problem which is a strictly convex quadratic programming (QP).
Considering its low-dimensional yet multi-constrained nature, we generalize Seidel's method~\cite{Seidel1991SmallDLP}, which is originally designed for solving linear programming with linear complexity, to handle with this minimum-norm.
Compared with several general-purpose solvers~\cite{ferreau2014qpoases,osqp,frison2020hpipm} for QP, this method achieves the capability of obtaining analytical solution within a significantly shorter time.

MVIE, known as one of the most challenging extremal ellipsoid problems~\cite{Khachiyan1993ComplexityMVIE}, is highly demanded in a number of applications~\cite{lovasz1986algorithmic,lenstra1983integer}.
\rv{Most existing methods adopt semidefinite programming (SDP) formulations~\cite{Ben2001LecturesMCO,Deits2015ComputingIRIS} with interior point method variants for solution~\cite{Khachiyan1993ComplexityMVIE,Zhang2003NumericalMVIE,Nesterov1994InteriorPMCP,MosekOPT}.
However, they struggle with computational efficiency when handling massive constraint due to large equation systems required in each iteration.
To reduce computational overhead without sacrificing solution quality, we present an equivalent second-order conic programming (SOCP) formulation for MVIE, bringing about a noticeable boost in computational efficiency.}

Moreover, especially in 2-D scenarios, we gain ultra computational efficiency by proposing an analytical method for MVIE.
The method exhibits a time complexity linear in the number of the hyperplanes of the input convex polytope, which is given for the first time to the best of our knowledge.
As the dual problem of MVIE, Minimum Volume Enclosing Ellipse (MVEE) has long been equipped with linear-time analytical algorithms~\cite{welzl2005smallest} based on its LP-type problem structure~\cite{sharir1992combinatorial}. 
\rv{However, the corresponding approach for MVIE remained absent for decades due to failing LP-type problem {properties} and lacking analytical solutions for {basis computation}\cite{matouvsek1992subexponential}.
We address these challenges through an improved randomized algorithm with problem reformulation and a bottom-up strategy inspired by GJK algorithm's distance subalgorithm~\cite{montanari2017improving}.}
Consequently, the specialized 2-D method achieves a substantial efficiency gain compared to other state-of-the-art methods~\cite{Deits2015ComputingIRIS,MosekOPT}, surpassing them by several orders of magnitude.

\rv{Building upon the above methods for RsI and MVIE, FIRI generates high-\textit{quality} convex polytopes with \textit{manageability} while maintaining high \textit{efficiency}.
To evaluate the performance of FIRI in these three requirements, we conduct comprehensive comparisons with various polytope generation algorithms~\cite{Deits2015ComputingIRIS,Liu2017PlanningDF,Zhong2020GeneratingLCP}.
The results provide compelling evidence that FIRI outperforms other approaches across all three requirements.}
Moreover, we extensively perform real-world applications to showcase the applicability of FIRI, which involves a 2-D vehicle with non-holonomic constraint and a 3-D quadrotor, as well as point-mass and whole-body model.

In summary, the contributions are:
\begin{itemize}
    \item [1)] \rv{RsI is introduced to ensure the manageability of the generated polytope. For its few-variable but rich-constraint nature, a specialized and efficient solver is designed.}
    \item [2)] \rv{A novel SOCP formulation for MVIE is proposed, avoiding the positive definite constraints and improving the computational efficiency.}
    \item [3)] Especially for 2-D MVIE, a linear-time analytical algorithm is introduced for the first time, further enabling ultra-fast computational performance.
    \item [4)] Building upon the above methods, a novel convex polytope generation algorithm FIRI is proposed. Extensive experiments verify its superior comprehensive performance in terms of quality, efficiency, and manageability. 
\end{itemize}

\section{Related Work}
\label{sec::Related Work}
\subsection{Generating Free Convex Polytope}

Deits et al.~\cite{Deits2015ComputingIRIS} propose Iterative Region Inflation by Semidefinite programming (IRIS) with the objective of generating the largest possible free convex polytope.
IRIS involves inflating an ellipsoid, using it as a seed, to obtain a convex polytope formed by the intersection of contact planes. 
Then in the next iteration, the MVIE of the obtained convex polytope is selected as the new seed for inflation.
However, this method requires solving semidefinite programming problems in each iteration to obtain the MVIE, which significantly hampers the computational efficiency.
\rv{
Furthermore, the hyperplane computation of IRIS cannot directly accommodate additional seed containment constraint, lacking manageability.
In contrast, our RsI for FIRI unifies the forms of obstacle exclusion and seed containment constraints, enabling simpler and more efficient solutions.}
\rv{Wu et al.~\cite{wu2024optimal} use an IRIS-like method to obtain convex polytope through ellipsoid-based hyperplanes.
They propose an ADMM-based approach that iteratively refines polytopes and optimizes trajectories to achieve shorter and better trajectory.}
Based on IRIS, Dai et al.~\cite{dai2023certified} propose C-IRIS, which focuses on mapping collision in the task space to the configuration space based on kinematics and then generates certified safe convex hulls in the configuration space. 
\rv{Similarly, Mark et al.\cite{petersen2023growing} propose IRIS-NP, extending IRIS to configuration space via nonlinear programming.
Subsequently, building upon IRIS-NP, Werner et al.\cite{werner2024faster} significantly accelerate polytope generation by sampling nearby configuration-space obstacles.
The above IRIS derivatives primarily target multi-joint robotic arms in configuration space, differing from our focus.}
To address the efficiency and local manageability deficiencies of IRIS, Liu et al.~\cite{Liu2017PlanningDF} propose the Regional Inflation by Line Search algorithm (RILS) which takes line segment as the seed input.  
RILS first generates a maximal ellipsoid that contains the segment yet excludes all obstacles.
\rv{Then RILS inflates the ellipsoid to form a convex polytope with contact planes from the obstacles, which fundamentally aligns with the inflation step in a single iteration of IRIS.}
RILS shows high computational speed. 
Yet using input line as major axis of the ellipsoid, it tends to produce conservative convex hull.
Specifically for voxel maps, Gao et al.~\cite{Gao2020TeachRepeatReplanAC} develop Parallel Convex Cluster Inflation algorithm. 
Starting from an unoccupied seed voxel, it grows incrementally in layers along coordinate axis using visibility, maintaining voxel set convexity. 
Despite parallel computing acceleration, this algorithm reaches near real-time only at low map resolution.

On the other hand, Savin et al.~\cite{savin2017algorithm} utilize the concept of space inversion. 
\rv{Using spherical polar mapping from the input seed point, they flip all the obstacle points outside. 
After computing convex hull of the inverted obstacle points, they transform it back to the original space for the final hull. 
The strong nonlinearity of spherical polar mapping limits this method, as volume loss in inverted space creates gaps between final polytope and obstacles, yielding conservative results.}
To address this limitation, Zhong et al.~\cite{Zhong2020GeneratingLCP} adopt sphere flipping mapping. 
Due to the properties of the hidden point remove operator inherent~\cite{katz2007direct} in this mapping, this approach obtains the visible star-convex region around the seed. 
\rv{They then partition this star-convex region into final convex polytope through a heuristic method. 
Yet this approach uses Quickhull~\cite{barber1996quickhull}, which degrades with points distributed near the sphere after mapping. 
In addition, the heuristic partitioning yields conservative result.}

Conclusively, existing methods consistently suffer from at least one of the following issues: inefficiency, conservatism, and lack of manageability, thereby limiting their practicality. 

\subsection{Maximum Volume Inscribed Ellipsoid~(MVIE)}
MVIE is also known as inner Löwner-John ellipsoid~\cite{john2014extremum}. 
Nesterov et al.~\cite{Nesterov1994InteriorPMCP} utilize an interior-point algorithm with a specialized rescaling method on each iteration to achieve a polynomial time solution of MVIE, \rv{surpassing} ellipsoid algorithm~\cite{tarasov1988method}.
Khachiyan et al.~\cite{khachiyan1990complexity} transform the problem into a sequence of subproblems with only linear constraints, constructed by using the barrier method. 
This approach requires fewer computations compared to Nesterov's method~\cite{Nesterov1994InteriorPMCP}.
Then Anstreicher et al.~\cite{Anstreicher2002ImprovedMVIE} make improvements to both methods~\cite{Nesterov1994InteriorPMCP,khachiyan1990complexity} and demonstrate that computing an approximate analytic center of the polytope beforehand can reduce the complexity effectively.
Zhang et al.~\cite{Zhang2003NumericalMVIE} also provide a modification of Khachiyan's approach~\cite{khachiyan1990complexity} by replacing the inefficient primal barrier function method with a primal-dual interior-point method to solve the subproblems.
Additionally, instead of dealing with a number of subproblems, they propose a novel primal-dual interior-point method free of matrix variables to solve MVIE directly.
Through nonlinear transformations, they eliminate the positive-definite constraint on the coefficient matrix of the ellipsoid during the iteration process and provide a proof demonstrating that these nonlinear transformations preserve the uniqueness of the solution.
Building upon similar idea of eliminating matrix variables, Nemirovskii~\cite{nemirovski1999self} reformulates MVIE as a saddle-point problem using Lagrangian duality.
\rv{They then use path-following method for finding approximate saddle points based on Nesterov's self-concordance theory~\cite{Nesterov1994InteriorPMCP}.
However, these interior-point methods struggle with scenarios having constraints far exceeding space dimension.}
In these scenarios which are the focus of this paper, the large-scale system of linear equations that these methods need to solve at each iteration prevents them from computing MVIE within an acceptable time.
To address inefficiency, Lin et al.~\cite{lin2018maximum} employ the fast proximal gradient method~\cite{beck2009fast} to introduce a first-order optimization-based approach for MVIE.
\rv{Although this approach significantly improves efficiency, it lacks exact solutions due to approximating non-differentiable indicator with an one-sided Huber function for positive-definite constraints.}

MVIE is the most challenging problem among the extremal ellipsoid problems~\cite{Khachiyan1993ComplexityMVIE}. 
Other extremal ellipsoid problems, such as Minimum Volume Enclosing Ellipsoid (MVEE), can be transformed into MVIE with a linear reduction, which is irreversible.
Beyond interior-point methods, analytical solutions of MVEE can be obtained using a linear-time randomized method in 2-D case~\cite{welzl2005smallest}.
However, for the more challenging MVIE, the corresponding linear complexity algorithm are absent for several decades. This paper will address this gap.

\section{Fast Iterative Region Inflation}

\subsection{Problem Formulation}
\label{sec::Problem}
Consider a convex seed in an obstacle-rich $n$-dimensional environment, where $n\in\{2,3\}$. 
The geometrical shape of the seed is given by the $\mathcal{V}$-representation~\cite{Toth2017HandbookDCG} of a convex polytope, i.e., convex combination of a finite number of points
\begin{equation}
\mathcal{Q}=\conv\cbrac{v_1, \dots, v_s},
\end{equation}
in which $v_1, \dots, v_s\in\mathbb{R}^n$ are allowed to be redundant, which means that they are not required to only contain the extreme points of $\mathcal{Q}$. 
The obstacle region $\mathcal{O}$ is, albeit nonconvex, assumed to be the union of convex obstacles $\mathcal{O}=\cup_{i=1}^{N}\mathcal{O}_i$, of which the $i$-th one is determined by $s_i$ points
\begin{equation}
\mathcal{O}_i=\conv\cbrac{u_{i,1}, \dots, u_{i,s_i}}.
\end{equation}
\rv{which additionally enables FIRI to directly process polytope-type obstacles, as opposed to point-only methods~\cite{Liu2017PlanningDF,Zhong2020GeneratingLCP} that require discretization to handle such obstacles.}
We require no collision between the seed $\mathcal{Q}$ and the obstacle region $\mathcal{O}$, thus implying  $\mathcal{Q}\cap\mathcal{O}=\varnothing$.

Our problem is to compute an obstacle-free convex polytope $\mathcal{P}$ which is required to contain the seed $\mathcal{Q}$ while excluding all obstacles $\mathcal{O}$. 
Besides, $\mathcal{P}$ should have the largest possible volume within a prescribed region of interest.
For convenience, we define that the boundaries of the prescribed region are considered as obstacles and are encoded into $\mathcal{O}$, which makes the volume of obstacle-free space surrounding $\mathcal{Q}$ bounded.
Concluding above requirements yields the optimization
\begin{equation}
    \label{eq:MaxVolFreePolytope}
    \max_{\mathcal{P}} ~{\vol\rbrac{\mathcal{P}}},~~\mathrm{s.t.}~\mathcal{Q}\subseteq\mathcal{P},~\rv{\mathcal{O}\cap \text{int}\rbrac{\mathcal{P}}=\varnothing},
    \end{equation}
    where $\vol\rbrac{\mathcal{P}}$ denotes the volume of the convex polytope $\mathcal{P}$ and \rv{$\text{int}\rbrac{\cdot}$ denotes the interior of a set}.
    Note that the solution set of (\ref{eq:MaxVolFreePolytope}) will never be empty since the seed $\mathcal{Q}$ itself is already a feasible solution.

\begin{figure*}[ht]
	\centering
	\includegraphics[width=0.95\linewidth]{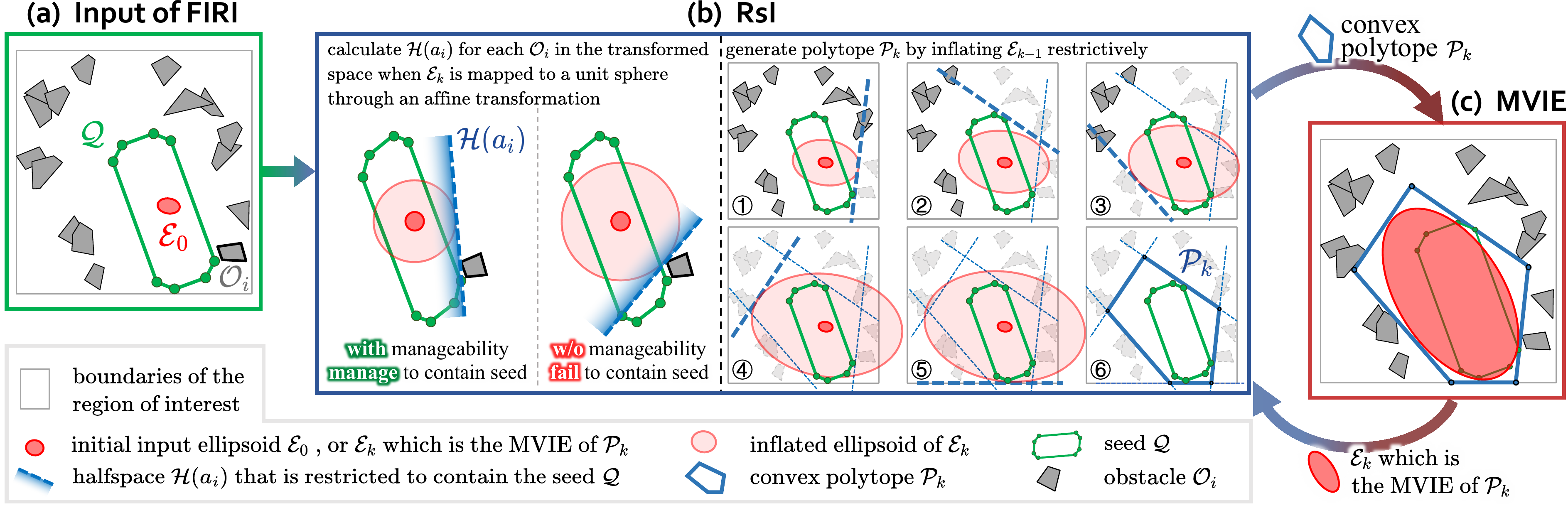}
	\vspace{-0.3cm}
	\caption{
		Overview of the computation process of FIRI, corresponding to the iterative modules RsI and MVIE calculation depicted in Algorithm.~\ref{alg:FIRI}. 
        \rv{The left diagram of (b) illustrates a specific example of halfspace computation w.r.t. $\mathcal{O}_i$ in the transformed space, comparing scenarios with and without manageability.}
        \raisebox{.5pt}{\textcircled{\raisebox{-.9pt} {1}}}-\raisebox{.5pt}{\textcircled{\raisebox{-.9pt} {6}}} of (b) provide a visualization combined the process in Line~\ref{algl:second_start}-\ref{algl:BackTransformHPolytope}. 
        The increasing size of the inflated ellipsoid in (b) corresponds to the iterative search for the nearest halfspace in Line~\ref{algl:second_start}.
        The increasing number of halfspaces corresponds to Line~\ref{algl:IntersectNewHalfspace}, and the decreasing obstacles correspond to Line~\ref{algl:second_end}.
	}
	\label{fig:FIRI}
	\vspace{-0.4cm}
\end{figure*}

\subsection{Algorithm Overview}
\label{sec::FIRI}
Even if we simplify the optimization~(\ref{eq:MaxVolFreePolytope}) by disregarding the constraint of including the seed, specifically $\mathcal{Q}\subseteq\mathcal{P}$, it will be transformed into \hidden{a generalization of the challenging \textit{potato peeling} problem} a challenging problem, \rv{requiring $O(N^7)$~\cite{Chang1986PolySolPPP} in 2-D and NP-hard~\cite{song2016cofifab} in 3-D.} 
Not to mention that bringing in this constraint \rv{further }complicates the optimization~(\ref{eq:MaxVolFreePolytope})\hidden{ even more than \textit{potato peeling}}.
\rv{For efficiency-quality balance}, we only seek a high-quality feasible solution to (\ref{eq:MaxVolFreePolytope}) efficiently instead of the global optimal solution.
Additionally, since the computation of the objective function vol$(\mathcal{P})$ of (\ref{eq:MaxVolFreePolytope}) is at least as hard as an NP-complete problem~\cite{dyer1988complexity},
we optimize a reasonable lower bound of vol$(\mathcal{P})$, which is straightforward to evaluate, to maximize the original objective function.
As adopted in~\cite{Deits2015ComputingIRIS}, we also choose the volume of the MVIE of $\mathcal{P}$ as this lower bound. 

To efficiently maximize the lower bound of the volume of the convex polytope $\mathcal{P}$ while satisfying the constraints of the original problem~(\ref{eq:MaxVolFreePolytope}), we propose a novel algorithm called \textbf{F}ast \textbf{I}terative \textbf{R}egion \textbf{I}nflation (FIRI) as shown in Algorithm~\ref{alg:FIRI}.
Any ellipsoid that is strictly contained in $\mathcal{Q}$ can be used to initialize the algorithm.
FIRI iteratively executes two modules: RsI and MVIE within its outer loop (Line~\ref{algl:FIRI_start}-\ref{algl:EllipsoidProgramming}). 
In the $k$-th iteration, the former module takes an ellipsoid as input and expands the ellipsoid to obtain a new convex polytope $\mathcal{P}_k$, and the latter module takes the convex polytope $\mathcal{P}_k$ as input and computes its MVIE $\mathcal{E}_k$.
As illustrated in Fig.~\ref{fig:FIRI}, the output of these two modules serves as the input for each other.
Then we provide the details of these two modules and the convergence and manageability of FIRI.

In this paper, we define an ellipsoid $\mathcal{E}$ by 
\begin{equation}
    \label{eq:ellipsoid}
    \mathcal{E}=\cBrac{p ~\big| ~p=A_\mathcal{E}D_\mathcal{E}x+b_\mathcal{E},~x\in\mathbb{R}^n,~\norm{x}=1},
\end{equation}
where $A_\mathcal{E}\in\mathbb{R}^{n\times n}$ is orthonormal, $D_\mathcal{E}\in\mathbb{R}^{n\times n}$  is diagonal and positive-definite, and $b_\mathcal{E}\in\mathbb{R}^n$. 
The diagonal elements of $D_\mathcal{E}$ correspond to the lengths of the semi-axis of $\mathcal{E}$.

\setlength{\textfloatsep}{2pt}
\begin{algorithm}[t]
    \label{alg:FIRI}
    \DontPrintSemicolon
    \LinesNumbered
    \SetAlgoLined
    \caption{\textbf{FIRI}}
    \textbf{Notion:} ~number of obstacles $N$,\;
    ~~~~~~~~~~~parameters of ellipsoid $\mathcal{E}$:~$A_\mathcal{E},D_\mathcal{E},b_\mathcal{E}$,\;
    ~~~~~~~~~~~halpspace $\mathcal{H}\rbrac{a}$ defined in (\ref{eq:halfspace})\;
    \KwIn{~~seed $\mathcal{Q}$, obstacles $\mathcal{O}$, initial ellipsoid $\mathcal{E}_0$,\;  ~~~~~~~~~~~threshold $\rho$}
    \KwOut{convex polytope $\mathcal{P}$}
    \Begin
    {
        $k\leftarrow0$\;
        \Repeat{$\vol\rbrac{\mathcal{E}_k}\leq(1+\rho)\vol\rbrac{\mathcal{E}_{k-1}}$}
        {
            $k\leftarrow k+1$\label{algl:FIRI_start}\;
            $\textcolor{blue}{\text{/* RsI starts */}}$\;
            $\mathcal{I}\leftarrow\cbrac{1,\dots,N}$\;
            $\mathcal{E}\leftarrow \mathcal{E}_{k-1}, ~\bar{\mathcal{P}}\leftarrow\mathbb{R}^n$\;
            $\bar{\mathcal{Q}}\leftarrow D_\mathcal{E}^{-1}A_\mathcal{E}\tp\rbrac{\mathcal{Q}-b_\mathcal{E}}$\label{algl:TransformPolytope}\;
            \ForEach{$i\in\mathcal{I}$}
            {
                $\bar{\mathcal{O}}_i\leftarrow D_\mathcal{E}^{-1}A_\mathcal{E}\tp\rbrac{\mathcal{O}_i-b_\mathcal{E}}$\label{algl:TransformObstacle}\;
                $a_i\leftarrow\argmax_{a\in\mathbb{R}^n}{a\tp a},$\label{algl:RestrictiveInflationStart}
                $~~~~~~~\mathrm{s.t.}~\bar{\mathcal{Q}}\subseteq\mathcal{H}\rbrac{a},~\rv{\bar{\mathcal{O}}_{i}\cap{\text{int}\rbrac{\mathcal{H}\rbrac{a}}}=\varnothing}$\;
            }
            \Repeat{$\mathcal{I}=\varnothing$}
            {
                $j\leftarrow\argmin_{i\in\mathcal{I}}{a_i\tp a_i}$\label{algl:second_start}\;
                $\bar{\mathcal{P}}\leftarrow\bar{\mathcal{P}}\cap\mathcal{H}\rbrac{a_j}$\label{algl:IntersectNewHalfspace}\;
                $\mathcal{I}\leftarrow\mathcal{I}\backslash\cBrac{i\in\mathcal{I}~\big|~\rv{\bar{\mathcal{O}}_i\cap{\text{int}\rbrac{\mathcal{H}\rbrac{a_j}}}=\varnothing}}$\label{algl:second_end}\;
            }
            $\mathcal{P}_k\leftarrow A_\mathcal{E}D_\mathcal{E}\bar{\mathcal{P}}+b_\mathcal{E}$\label{algl:BackTransformHPolytope}\;
            $\textcolor{blue}{\text{/* RsI ends */}}$\;
            $\textcolor{blue}{\text{/* MVIE starts */}}$\;
            $\mathcal{E}_k\leftarrow\mathcal{E}^*\rbrac{\mathcal{P}_k}$\label{algl:EllipsoidProgramming}\;
            $\textcolor{blue}{\text{/* MVIE ends */}}$\;
        }\label{algl:FIRI_end}
        \Return{$\mathcal{P}_k$}
    }
\end{algorithm}

\subsubsection{RsI}
For each obstacle $\mathcal{O}_i$, we maximally inflate the input ellipsoid $\mathcal{E}$ under the constraint that there exists a halfspace which does not contain the obstacle yet includes both the inflated ellipsoid and seed $\mathcal{Q}$.
Then we use these halfspaces to forms a new polytope as the output of RsI.

Specifically, $\mathcal{Q}$ and $\mathcal{O}_i$ are transformed (Line~\ref{algl:TransformPolytope} and~\ref{algl:TransformObstacle}) by the inverse affine map determined by the ellipsoid $\mathcal{E}$ generated in the previous iteration. Since $\mathcal{Q}$ and $\mathcal{O}_i$ are both in $\mathcal{V}$-representation, their images after transformation are still the convex combinations of the images of their vertices, i.e.,
\begin{align}
\bar{\mathcal{Q}}&=\conv\cbrac{D_\mathcal{E}^{-1}A_\mathcal{E}\tp\rbrac{v_j-b_\mathcal{E}},~1\leq j\leq s},\\
\bar{\mathcal{O}_i}&=\conv\cbrac{D_\mathcal{E}^{-1}A_\mathcal{E}\tp\rbrac{u_{i,j}-b_\mathcal{E}},~1\leq j\leq s_i}.
\end{align}
It is important that in the transformed space the ellipsoid $\mathcal{E}$ is transformed to a unit ball $\mathcal{B}$ as
\begin{equation}
    \mathcal{B}=\cBrac{x\in\mathbb{R}^n~|~\norm{x}=1},
\end{equation}
whose collision check is far cheaper to handle than ellipsoids.

Then we calculate a restrictive halfspace for each transformed obstacle $\bar{\mathcal{O}}_i$ in the first inner loop (Line~\ref{algl:TransformObstacle}-\ref{algl:RestrictiveInflationStart}).
Compared to the methods that lack the manageability to include the seed~\cite{Deits2015ComputingIRIS,Liu2017PlanningDF}, the proposed RsI requires that the halfspace contains $\bar{\mathcal{Q}}$ and the corresponding inflated ball $\mathcal{B}_i$ yet excludes $\bar{\mathcal{O}}_i$.
Additionally, it aims to maximize the size of $\mathcal{B}_i$ which is defined by
\begin{equation}
    \mathcal{B}_i=\cBrac{x\in\mathbb{R}^n~|~\norm{x}=a_i\tp a_i},
\end{equation}
\rv{where $a_i$ is the only contact point of the halfspace boundary with the inflated ball $\mathcal{B}_i$.}
The halfspace $\mathcal{H}\rbrac{a_i}$ is defined as
\begin{equation}
    \label{eq:halfspace}
    \mathcal{H}\rbrac{a_i}=\cBrac{x\in\mathbb{R}^n~\big|~a_i\tp x\leq a_i\tp a_i}.
\end{equation}
The calculation of the restrictive halfspace can be written as an optimization as shown in Line~\ref{algl:RestrictiveInflationStart}, for which we provide an efficient solver in Sec.~\ref{sec:Solving_Restrictive_Inflation}.
In the left diagram of Fig.~\ref{fig:FIRI}(b), we provide a specific example of the restrictive halfspace computation in the transformed space, \rv{comparing scenarios with and without manageability.
Notably, in the scenario without manageability (employed by IRIS), the exclusive focus on maximizing ellipsoid inflation results in the computed halfspace being unable to guarantee seed point containment.
}

In the second inner loop (Line~\ref{algl:second_start}-\ref{algl:second_end}), 
we generate a new convex polytope based on the obtained halfspaces.
We iteratively finds the closest halfspace $\mathcal{H}\rbrac{a_i}$, adds it into $\bar{\mathcal{P}}$, and then remove the halfspaces corresponding to the obstacles outside of $\mathcal{H}\rbrac{a_i}$. 
\rv{This widely adopted greedy strategy~\cite{Deits2015ComputingIRIS,Liu2017PlanningDF,werner2024faster} is effective in reducing the number of halfspaces that constitute the generated convex polytope, aiming to minimize  them as much as possible, compared to the large number of obstacles.}
Until all halfspaces are processed, an obstacle-free polytope in $\mathcal{H}$-representation~\cite{Toth2017HandbookDCG} can be formed in the transformed space. 
Then a new polytope $\mathcal{P}_k$ is obtained by recovering $\bar{\mathcal{P}}$ to the original space (Line~\ref{algl:BackTransformHPolytope}).
Thus, the output of FIRI is an obstacle-free convex polytope in $\mathcal{H}$-representation, i.e., intersection of $m$ halfspaces
\begin{equation}
\label{eq:convex_polytope}
\mathcal{P}=\cBrac{x\in\mathbb{R}^n~\big|~A_\mathcal{P}x\leq b_\mathcal{P}},
\end{equation}
where $A_\mathcal{P}\in\mathbb{R}^{m\times n}$ and $b_\mathcal{P}\in\mathbb{R}^m$. 
As shown in \raisebox{.5pt}{\textcircled{\raisebox{-.9pt} {1}}}-\raisebox{.5pt}{\textcircled{\raisebox{-.9pt} {6}}} of Fig.~\ref{fig:FIRI}(b), we present a combined representation of the two processes mentioned above (Line~\ref{algl:second_start}-\ref{algl:BackTransformHPolytope}).

\subsubsection{MVIE}
As long as a closed $\mathcal{P}$ has nonempty interior, its unique MVIE~\cite{agarwal2015duality} $\mathcal{E}^*\rbrac{\mathcal{P}}$ can be determined by solving 
\begin{equation}
\label{eq:AbstractMVIE}
\max_{A_\mathcal{E},D_\mathcal{E},b_\mathcal{E}}{\vol\rbrac{\mathcal{E}}},~\mathrm{s.t.}~\mathcal{E}\subseteq\mathcal{P},
\end{equation}
which is employed in Line~\ref{algl:EllipsoidProgramming}.
\rv{Notably, MVIE takes exclusively one convex polytope as input, independent of any obstacles used to generate the input polytope.}
To solve MVIE, we propose efficient methods in Sec.~\ref{sec:SOCP_MVIE} and Sec.~\ref{sec:exact_MVIE} so that the computational overhead of MVIE will no longer be a stumbling block that prevents this monotonically inflating MVIE approach from being applied to real-time scenarios~\cite{Deits2015ComputingIRIS}.

\subsubsection{Manageability and Convergence of FIRI}
During the iterative computation, FIRI always ensures the feasibility of the solution, which means the generated convex polytope $\mathcal{P}_k$ maintains satisfacting the constraints of the original problem as $k$ increases.
Moreover, it maintains the monotonicity of the volume of MVIE $\mathcal{E}_k$, which is the lower bound of the volume of the convex polytope $\mathcal{P}_k$.
Then by analyzing the feasibility and monotonicity of the output of FIRI, we explain the manageability and convergence they bring about.

For the feasibility, since each halfspace computed in RsI (Line~\ref{algl:RestrictiveInflationStart}) satisfies the original constraints, the new convex polytope $\mathcal{P}_{k}$ formed by the intersection of these halfspaces is guaranteed to be feasible :
\begin{equation}
\mathcal{Q}\subseteq\mathcal{P}_{k},~\rv{\mathcal{O}\cap\text{int}\rbrac{\mathcal{P}_{k}}=\varnothing}
\end{equation}
which gives the manageability of FIRI.
Additionally, as defined in Sec~\ref{sec::Problem}, the prescribed region of interest is bounded, whose boundaries are encoded into $\mathcal{O}$, thus the feasibility also indicates that RsI always generates a closed polytope $\mathcal{P}_{k}$.

For the monotonicity, in the $k$-th iteration, we inflate the unit ball $\mathcal{B}$ for each transformed obstacle $\bar{\mathcal{O}}_i$ in RsI, which always ensures the contact point $a_i$ satisfies $\norm{a_i}\geq1$. 
Thus, in the transformed space, $\bar{\mathcal{P}}$ acquired from the second inner loop (Line~\ref{algl:second_start}-\ref{algl:second_end}) always holds $\mathcal{B}\subseteq\bar{\mathcal{P}}$, which means that in the original space (Line~\ref{algl:BackTransformHPolytope}) we have $\mathcal{E}_{k-1}\subseteq\mathcal{P}_k$.
As $\mathcal{E}_{k}$ is the largest ellipsoid contained by $\mathcal{P}_k$, we have
\begin{equation}
    \vol\rbrac{\mathcal{E}_{k-1}}\leq \vol\rbrac{\mathcal{E}_{k}}.
\end{equation}
which gives the monotonicity.

Then a sequence $\cbrac{\mathcal{P}_1,\mathcal{E}_1,\dots,\mathcal{P}_k,\mathcal{E}_k,\dots}$ is generated by repeating the iteration. 
Since $\vol\rbrac{\mathcal{E}_k}$ is non-decreasing and bounded, the ellipsoid volume will converge to a finite value.
Consequently, FIRI terminates when this lower bound for $\vol\rbrac{\mathcal{P}_k}$ cannot be sufficiently improved (Line~\ref{algl:FIRI_end}).

In conclusion, RsI brings manageability to FIRI, \rv{while monotonic iterative updates optimize the polytope volume's lower bound for a high-quality output.
Notably, the efficiency of FIRI relies strongly on the performance of solving two optimizations in Line~\ref{algl:RestrictiveInflationStart} and~\ref{algl:EllipsoidProgramming}.
In subsequent sections, we propose efficient and reliable subalgorithms exploiting the geometric structure of these optimizations, greatly benefiting the computational efficiency of FIRI.}

\section{Solving Restrictive Halfspace Computation in RsI via SDMN}
\label{sec:Solving_Restrictive_Inflation}
\subsection{Reformulation of Restrictive Halfspace Computation}
In this section, we focus on designing efficient method for the computation of halfspaces in RsI, defined in Line~\ref{algl:RestrictiveInflationStart} of Algorithm~\ref{alg:FIRI}.
Combining the definition of the halfspace $\mathcal{H}\rbrac{a_i}$ in (\ref{eq:halfspace}), we formulate the halfspace calculation into
\begin{subequations}
    \label{eq:RestrictiveInflationNonConvex}
    \begin{align}
    \max_{a_i\in\mathbb{R}^n}&{~a_i\tp a_i},\\
    \mathrm{s.t.}~&~v\tp a_i\leq a_i\tp a_i,~\forall v\in\bar{\mathcal{Q}},\\
    &~u\tp a_i\geq a_i\tp a_i,~\forall u\in\bar{\mathcal{O}}_i.
    \end{align}
\end{subequations}
Algorithm~\ref{alg:FIRI} keeps $a_i\tp a_i>0$, which ensures that the origin always lies within the interior of the halfspace $\mathcal{H}\rbrac{a_i}$. 
Thus, although such a maximization of the inflated ball $\mathcal{B}_i$ is nonconvex, we can obtain its equivalent minimum-norm formulation through its polar duality. 
Specifically, we \rv{reformulate the problem with a new variable $b$} \hidden{directly optimize the polar dual vector of $a_i$} by substituting $a_i=b/(b\tp b)$ and obtain an equivalent $L_2$-norm minimization
\begin{subequations}
    \label{eq:RestrictiveInflationQP}
    \begin{align}
    \min_{b\in\mathbb{R}^n}&{~b\tp b},\\
    \mathrm{s.t.}~&~v\tp b\leq 1,~\forall v\in\bar{\mathcal{Q}},\\
    &~u\tp b\geq 1,~\forall u\in\bar{\mathcal{O}}_i,
    \end{align}
\end{subequations}
which has a low-dimension but multi-constraint nature.
\rv{In reformulation (\ref{eq:RestrictiveInflationQP}), both types of constraints share a unified inequality form. 
This formulation handles cases with and without the seed containment constraint, allowing subsequent solver SDMN adapt readily. 
In contrast, prior work such as IRIS~\cite{Deits2015ComputingIRIS} struggles to add seed containment while maintaining the original problem structure, making it challenging to achieve manageability via straightforward constraint addition.}

\subsection{Solution to Small-Dimensional Minimum-Norm with Massive Constraints}

For efficiently solving the new formulation~(\ref{eq:RestrictiveInflationQP}) of restrictive halfspace computation, we propose an analytical method for such \textbf{S}mall \textbf{D}imensional \textbf{M}inimum-\textbf{N}orm, called SDMN.
This method generalizes Seidel's randomized algorithm~\cite{Seidel1991SmallDLP} from Linear Programming~(LP) to minimum-norm, and enjoys complexity linear in the constraint number.

Without loss of generality, we consider the following general small-dimensional minimum-norm problem,
\begin{equation}
    \label{eq:exampleL2}
    \min_{y\in\mathbb{R}^n}{ y\tp y},~\mathrm{s.t.}~Ey\leq f,
\end{equation}
where $E\in\mathbb{R}^{d\times n}$, $f\in\mathbb{R}^{d}$, 
$d$ denotes the number of constraints, which is much larger than $n$.
Hereafter, we present a randomized algorithm with linear complexity for the small-dimensional minimum-norm (\ref{eq:exampleL2}) in Algorithm~\ref{alg:SDMN}.

As shown in Algorithm~\ref{alg:SDMN}, this is a recursive algorithm. 
In the following, we first provide an outline of the algorithm, then we give a detailed description of how to construct the recursive problem, and finally present  a complexity analysis.

\begin{algorithm}[t]
    \label{alg:SDMN}
    \DontPrintSemicolon
    \LinesNumbered
    \SetAlgoLined
    \caption{\textbf{SDMN}}
    \textbf{Notion:~}  constraints already checked $\mathcal{I}$,\;
    ~~~~~~~~~~~~input and output of the recursive call ${{\mathcal{H}_E}'},~y'$\ \;
    \KwIn{~~set of halfspace constraints $\mathcal{H}_E$}
    \KwOut{$y$}
    \Begin
    {
        $y\leftarrow \mathbf{0}$\label{alg2:init}\;
        \If{$\mathrm{dim}(\mathcal{H}_E)==1$}
        {
            $y\leftarrow \textbf{OneDimMinNorm}(\mathcal{H}_E)$\label{alg2:one dim check}\;
            \Return{$y$}
        }
        $\mathcal{I}\leftarrow \{\}$\;
        \ForEach{$h\in\mathcal{H}_E~~\mathrm{in~a~random~order}$\label{alg2:check}}
        {
            \If{$y\notin h$}
            {
                $\textcolor{blue}{\text{/* Detail in Sec.~\ref{sec::Recursive Problem Construction} */}}$\;
                $\{M,v,{\mathcal{H}_E}'\}\leftarrow \textbf{HouseholderProj}(\mathcal{I},h)$\label{alg2:transformation}\;
                $y'\leftarrow \textbf{SDMN}({{\mathcal{H}_E}'})$\label{alg2:recursive_call}\;
                $y\leftarrow My'+v$\label{alg2:orign_solution}\;
            }
            $\mathcal{I}\leftarrow \mathcal{I}\cup\{h\}$\label{alg2:add constraint}\;
        }
        \Return{$y$}
    }
\end{algorithm}

\subsubsection{Algorithm Outline}
\label{sec::SDMN Algorithm Outline}
\begin{figure}[!t]
	\centering
	\includegraphics[width=1\linewidth]{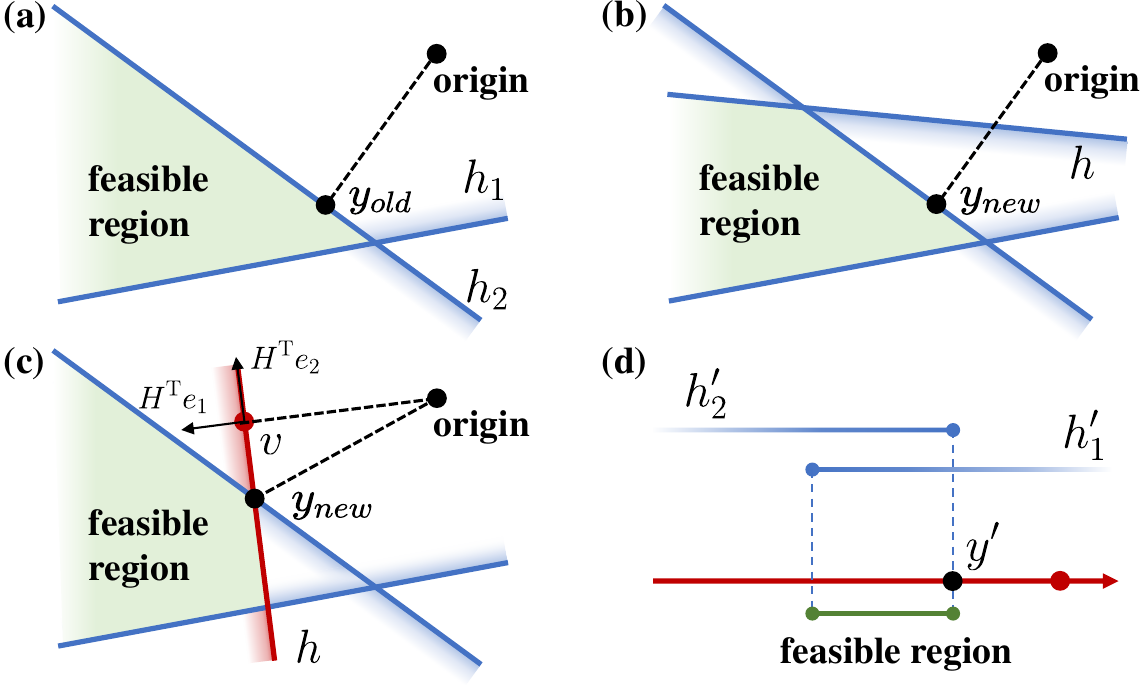}
	\vspace{-0.6cm}
	\caption{
		Illustration of a specific instance of Algorithm~\ref{alg:SDMN} in 2-D for Sec.~\ref{sec::SDMN Algorithm Outline} and Sec.~\ref{sec::Recursive Problem Construction}. 
        \textbf{For Sec.~\ref{sec::SDMN Algorithm Outline}}:
        \textbf{(a)}: Both the inequality constraints $h_1$ and $h_2$ have been checked, which means $\mathcal{I}=\{h_1, h_2\}$. 
        $y_{old}$ is the solution of the 2-D $L_2$-norm minimization under $\mathcal{I}$.
        \textbf{(a)}$\Rightarrow$\textbf{(b)}:~When $y_{old}$ does not violate the newly added constraint $h$, $y_{new}=y_{old}$.
        \textbf{(a)}$\Rightarrow$\textbf{(c)}:~When $y_{old}$ violates the new constraint $h$, we need to find a new solution $y_{new}$ on the constraint plane corresponding to $h$, implying equation constraint~(\ref{eq:exampleL2_active}).
        \textbf{For Sec.~\ref{sec::Recursive Problem Construction}}:
        The vector $v$ is normal to the constraint plane, and $H\tp e_1$, $H\tp e_2$ are a set of orthogonal basis of the 2-D space, where $H\tp e_2 \bot v$.
        \textbf{(d)}: We establish a new coordinate for the 1-D subspace on the constraint plane with $v$ as the origin and $H\tp e_2$ as the orthogonal basis, transform the checked constraints $h_1, h_2$ to this coordinate as $h'_1, h'_2$. Finally, we transforms (c) into a 1-D $L_2$-norm minimization with only inequality constraints (\ref{eq:exampleL2_active_n_1}).
	}
	\label{fig:SDMN}
\end{figure}

We denote $\mathcal{H}_E$ as the set of the hyperplanes corresponding to the constraints of the $L_2$-norm minimization (\ref{eq:exampleL2}), and denote $\mathcal{I}$ as the set of the constraints that have already been checked (Line~\ref{alg2:add constraint}).
Algorithm~\ref{alg:SDMN} starts with $y=\mathbf{0}$ which is the solution of an unconstrained $L_2$-norm minimization (Line~\ref{alg2:init}), and then gradually checks the constraint $h\in\mathcal{H}_E$ in a random order (Line~\ref{alg2:check}).
We check whether the solution under the constraints $\mathcal{I}$ violates the new constraint $h$. 
If it is not violated, the next constraint will be checked. 
If it is violated, $h$ must be active.
That is, the solution must be on the boundary of $h$, thus the minimization (\ref{eq:exampleL2}) under the constraints $\mathcal{I}\cup\{h\}$ can be written as
\begin{subequations}
    \label{eq:exampleL2_active}
    \begin{align}
    \min_{y\in\mathbb{R}^n}~&{ y\tp y},\\
    \mathrm{s.t.}~&~E_\mathcal{I} y\leq f_\mathcal{I}\label{eq:exampleL2_active iq},\\
    &~E_h y= f_h\label{eq:exampleL2_active eq},
    \end{align}
\end{subequations}
where $E_\mathcal{I}$ and $f_\mathcal{I}$ represent the coefficients corresponding to the hyperplanes in $\mathcal{I}$, $E_{h}$ and $f_{h}$ represent the coefficients of $h$.
Using the geometric structure of the problem~(\ref{eq:exampleL2_active}), we transform it into a subproblem of $(n-1)$ dimensional $L_2$-norm minimization with the same form as (\ref{eq:exampleL2}) in Line~\ref{alg2:transformation}, which is described in detail in Sec.~\ref{sec::Recursive Problem Construction}.
This transformation allows for a recursive call of Algorithm~\ref{alg:SDMN} (Line~\ref{alg2:recursive_call}).
Additionally, when $n=1$, the problem (\ref{eq:exampleL2}) is equivalent to a problem of computing the smallest absolute value in a interval (Line~\ref{alg2:one dim check}), whose solution can be calculated trivially. 
We assume that the subproblem can be successfully solved, and thus the new solution under the constraints $\mathcal{I}\cup\{h\}$ can be calculated (Line~\ref{alg2:orign_solution}).
Subsequently, the next constraint can be checked until all constraints in the set $\mathcal{H}_E$ are examined, from which we obtain the result of Algorithm~\ref{alg:SDMN}.
To provide an intuitive perception, we present an example of the violation check in Fig.~\ref{fig:SDMN}, where we use $y_{old}$ and $y_{new}$ to distinguish between solutions under $\mathcal{I}$ and $\mathcal{I}\cup\{h\}$.

\subsubsection{Recursive Problem Construction}
\label{sec::Recursive Problem Construction}
An essential process of Algorithm~\ref{alg:SDMN} is the recursive problem construction (Line~\ref{alg2:transformation}-\ref{alg2:recursive_call}), which transforms (\ref{eq:exampleL2_active}) with an equality constraint into a $(n-1)$ dimensional subproblem with the same form as the original problem (\ref{eq:exampleL2}).
\rv{This process's efficiency and stability greatly impact Algorithm~\ref{alg:SDMN}.}
Since $L_2$-norm is invariant under the orthogonal transformation, we establish a new $(n-1)$ dimensional Cartesian coordinate system on the constraint plane (\ref{eq:exampleL2_active eq}) to implement the construction of the reduced dimensional recursive problem. 
We provide a specific example of this process in Fig.~\ref{fig:SDMN}(c) and \ref{fig:SDMN}(d).
As illustrated in Fig.~\ref{fig:SDMN}(c), we take the point of $L_2$-norm minimization on the constraint plane as the origin of the coordinate system,
\begin{equation}
    \label{eq:origin of the coordinate system}
    v=\dfrac{f_h E_h\tp}{E_h E_h\tp}.
\end{equation}
It is obvious that any point $y$ on the constraint plane satisfies
\begin{equation}
    \label{eq:exampleL2_decomposition}
     y^\mathrm{T}y= (y-v)^\mathrm{T}(y-v)+ v^\mathrm{T}v,
\end{equation}
which means that, with the constraint (\ref{eq:exampleL2_active eq}), the $L_2$-norm minimization of $y$ is equivalent to the $L_2$-norm minimization of $(y-v)$ which can be transformed into an $(n-1)$ dimensional vector in the newly established coordinate system.

We construct the new coordinate system by finding an orthogonal matrix $H\in\mathbb{R}^{n\times n}$ that satisfies the requirement
\begin{equation}
    \label{eq:Householder reflection}
    Hv \propto e_j,
\end{equation}
where $e_j\in\mathbb{R}^n$ is an unit vector whose $j$-th element is 1.
We provide the details of the choice of $j$ and the computation of $H$ at the end of Sec.~\ref{sec::Recursive Problem Construction}.
The requirement (\ref{eq:Householder reflection}) indicates that the set of all column  vectors of orthogonal $H\tp$ except the $j$-th one form an orthonormal basis for the orthogonal complement of $v$, as shown in Fig.~\ref{fig:SDMN}(c).
Since $v$ is normal to the constraint plane, we adopt this orthonormal basis to define the new coordinate system of the $(n-1)$ dimensional subspace on the constraint plane, and introduce the corresponding coordinates $y'\in \mathbb{R}^{n-1}$.
We denote $M \in \mathbb{R}^{n\times(n-1)}$ as the matrix obtained by removing the $j$-th column of $H\tp$. Then $y$ in the origin coordinate system corresponding to $y'$ can be computed as
\begin{equation}
    \label{eq:exampleL2_decomposition2}
    y=My'+v.
\end{equation}
\rv{Based on the decomposition (\ref{eq:exampleL2_decomposition}) and (\ref{eq:exampleL2_decomposition2})}, minimizing the $L_2$-norm of $y$ on the constraint plane is equivalent to minimizing
\begin{equation}
    y^\mathrm{T}y - v^\mathrm{T}v={y'}^\mathrm{T}{M}\tp{M}{y'}={y'}^\mathrm{T}{y'}.
\end{equation}

Eventually, for (\ref{eq:exampleL2_active}) with an equality constraint, we construct its corresponding $(n-1)$ dimensional $L_2$-norm minimization with only inequality constraints on the constraint plane as
\begin{subequations}
    \label{eq:exampleL2_active_n_1}
    \begin{align}
    \min\limits_{{y'}\in\mathbb{R}^{n-1}}~&{y'}^\mathrm{T}{y'},\\
    \mathrm{s.t.}~&E_\mathcal{I}M{y'}\leq f_\mathcal{I}-E_\mathcal{I} v.
    \end{align}
\end{subequations}
As Fig.~\ref{fig:SDMN}(d) shows, the linear inequality constraints on ${y'}$ can be obtained by substituting (\ref{eq:exampleL2_decomposition2}) into the inequality constraint (\ref{eq:exampleL2_active iq}) of the original problem, and the set of their corresponding hyperplanes is denoted as ${\mathcal{H}_E}'$ (Line~\ref{alg2:transformation}-\ref{alg2:recursive_call}).
Obviously, (\ref{eq:exampleL2_active_n_1}) has the same structure as (\ref{eq:exampleL2}) and has a lower dimension, thus it can be solved by recursively calling Algorithm~\ref{alg:SDMN} (Line~\ref{alg2:recursive_call}), which gives rise to the formation of a recursive algorithmic structure.

Here we provide the details of $j$ and $H$ in (\ref{eq:Householder reflection}).
We obtain the orthogonal matrix $H$ by Householder reflection~\cite{householder1958unitary}. 
$v$ is a normal vector of the constraint plane proportional to the geometric scale of the problem, which has intuitive numerical stability. 
Thus, we use the normal vector $v$ to compute the Householder reflection. 
First, we set $j$
as the index of the element of $v$ with the largest absolute value as
\begin{equation}
    j=\mathrm{argmax}_{k\in\{1,...,n\}}\|v^\mathrm{T}e_k\|.
\end{equation}
Then the reflection vector $u$ that transforms $v$ to be parallel to $e_j$ can be calculated by
\begin{equation}
    u=v+\operatorname{sgn}(v_j)\|v\|e_j,
\end{equation}
whose corresponding Householder matrix can be calculated by
\begin{equation}
    \label{eq:Householder matrix}
    H=\mathbf{I}_n-\dfrac{2uu^\mathrm{T}}{u^\mathrm{T}u},
\end{equation}
which is an orthogonal matrix and satisfies the aforementioned requirement (\ref{eq:Householder reflection}).
The use of the obtuse reflection vector $u$ corresponding to the opposite direction of the largest absolute value element of $v$ prevents the Householder transformation from being ill-conditioned due to a small reflection angle. 
This procedure is implicitly equivalent to a single-step operation of the Householder QR factorization, which has higher numerical stability than the Gram-Schmidt orthogonalization which may fall victim to the catastrophic cancelation problem~\cite{Golub2013MatrixCMP}.

\subsubsection{Complexity Analysis}
To conclude this section, we give the complexity analysis of the randomized Algorithm~\ref{alg:SDMN}.
\rv{Each iteration may trigger recursive calls to $(n-1)$ dimensional subproblems. 
For small dimension $n$ with large constraint scale $d$, new constraints rarely become active, making recursive calls tiny.}
The expected complexity is $O(n!d)$~\cite{Seidel1991SmallDLP}, so that in the common dimensions $n\in\{2,3\}$, the expected complexity increases only linearly with the constraint scale\footnote{\rv{For specific derivation regarding the complexity, readers can refer to Theorem~1 of Seidel's work~\cite{Seidel1991SmallDLP}.}}. 
In addition, \rv{random constraint permutation preprocessing ensures performance independent of input order, nearly always achieving expected linear complexity.} 
With the help of this randomized algorithm, the complexity of handling all obstacles in the RsI of Algorithm~\ref{alg:FIRI} grows linearly only with the total number of vertices of the obstacles and the seed.

\section{Solving SOCP-Reformulation of MVIE via Affine Scaling Algorithm}
\label{sec:SOCP_MVIE}

\rv{In this section, we propose an efficient algorithm for MVIE with a low dimension but massive constraints, required in FIRI (Line~\ref{algl:EllipsoidProgramming} of Algorithm~\ref{alg:FIRI}). 
By carefully handling the orthogonal constraints on the elliptical matrix and the objective function involving the matrix determinant, we reformulate MVIE into a Second-Order Conic Programming (SOCP) form.
Using Affine Scaling~\cite{lagarias1990ii}, we solve this SOCP efficiently for MVIE.}

\subsection{SOCP-Reformulation of MVIE}
\label{sec:SOCP-Reformulation of MVIE}
To begin with, we concretize the definition of the original abstract MVIE problem~(\ref{eq:AbstractMVIE}).
According to the coefficients of the ellipsoid $\mathcal{E}$ defined in (\ref{eq:ellipsoid}), the diagonal elements of $D_\mathcal{E}$ are the lengths of the semi-axis of $\mathcal{E}$. 
Thus the objective $\vol\rbrac{\mathcal{E}}$ of the problem~(\ref{eq:AbstractMVIE}) is proportional to the determinant $\det\rbrac{D_\mathcal{E}}$~\cite{Boyd2004ConvexOPT}. 
Moreover, the semi-infinite constraint $\mathcal{E}\subseteq\mathcal{P}$ is equivalent to $\sbrac{\sbrac{A_\mathcal{P}A_\mathcal{E}D_\mathcal{E}}^2\mathbf{1}}^{\frac{1}{2}}\leq b_\mathcal{P}-A_\mathcal{P}b_\mathcal{E}$~\cite{Ben2001LecturesMCO} which is a second-order cone~(SOC) constraint, where $A_\mathcal{P}$, $b_\mathcal{P}$ are the coefficients of the halfspace constituting the convex polytope $\mathcal{P}$ defined in (\ref{eq:convex_polytope}), and $\sbrac{\cdot}$ implies entry-wise operations. 
Then the problem~(\ref{eq:AbstractMVIE}) can be written as
\begin{subequations}
\label{eq:OrthonormalityMVIE}
\begin{align}
\max_{A_\mathcal{E},D_\mathcal{E},b_\mathcal{E}}&{\det\rbrac{D_\mathcal{E}}},\label{eq::OrthonormalityMVIE_objective}\\
\mathrm{s.t.}~~&\sbrac{\sbrac{A_\mathcal{P}A_\mathcal{E}D_\mathcal{E}}^2\mathbf{1}}^{\frac{1}{2}}\leq b_\mathcal{P}-A_\mathcal{P}b_\mathcal{E},\label{eq::OrthonormalityMVIE_constraints}\\
&D_\mathcal{E}=\diag\rbrac{d_\mathcal{E}},~d_\mathcal{E}\in\mathbb{R}^n_{\geq0},\\
&A_\mathcal{E}\tp A_\mathcal{E}=I_n\label{eq::OrthonormalityMVIE_Orth},
\end{align}
\end{subequations}
where $\diag\rbrac{\cdot}$ indicates either constructing a diagonal matrix or taking all diagonal entries of a square matrix. However, this program still imposes orthonormality constraints on $A_\mathcal{E}$.
It is worth mentioning that if we use $A_\mathcal{E}D_\mathcal{E}$ as the decision variable, (\ref{eq:OrthonormalityMVIE}) becomes an SDP with a challenging orthogonal constraint, which is often used to solve MVIE~\cite{Deits2015ComputingIRIS,Ben2001LecturesMCO,MosekOPT}.

Then we eliminate the orthogonal constraints~(\ref{eq::OrthonormalityMVIE_Orth}).
Noting that  $A_\mathcal{E}D_\mathcal{E}^2A_\mathcal{E}\tp$ is always positive definite for the optimal solution of the non-degenerate problem (\ref{eq:OrthonormalityMVIE}).
Thus the Cholesky factorization  $A_\mathcal{E}D_\mathcal{E}^2A_\mathcal{E}\tp=L_\mathcal{E}L_\mathcal{E}\tp$ is unique~\cite{Horn2012MatrixA} for this solution, where $L_\mathcal{E}$ is a lower triangular matrix with positive diagonal entries. If we treat $L_\mathcal{E}$ as decision variables, since  $A_\mathcal{E}$ is orthonormal, the objective~(\ref{eq::OrthonormalityMVIE_objective}) and constraints~(\ref{eq::OrthonormalityMVIE_constraints}) can be written as
\begin{gather}
\det\rbrac{D_\mathcal{E}}=\sqrt{\det\rbrac{A_\mathcal{E}D_\mathcal{E}^2A_\mathcal{E}\tp}}=\det\rbrac{L_\mathcal{E}},\\
\sbrac{\sbrac{A_\mathcal{P}A_\mathcal{E}D_\mathcal{E}}^2\mathbf{1}}^{\frac{1}{2}}=\sbrac{\sbrac{A_\mathcal{P}L_\mathcal{E}}^2\mathbf{1}}^{\frac{1}{2}}.
\end{gather}
Consequently, the orthonormality constraint (\ref{eq::OrthonormalityMVIE_Orth}) on $A_\mathcal{E}$ is avoided, and (\ref{eq:OrthonormalityMVIE}) is equivalent to
\begin{subequations}
    \label{eq:NoOrthonormalityMVIE}
    \begin{align}
    \max_{L_\mathcal{E},b_\mathcal{E}}~&{\det\rbrac{L_\mathcal{E}}},\\
    \mathrm{s.t.}~~&\sbrac{\sbrac{A_\mathcal{P}L_\mathcal{E}}^2\mathbf{1}}^{\frac{1}{2}}\leq b_\mathcal{P}-A_\mathcal{P}b_\mathcal{E},\\
    &~L_\mathcal{E}~\mathrm{is~lower~triangular}.
    \end{align}
\end{subequations}

Now we simplify the objective function $\det\rbrac{L_\mathcal{E}}$ which is the product of the diagonal entries. 
Denoting the hypograph of geometric mean by
\begin{equation} \mathcal{K}_{1/n}=\cBrac{\rbrac{x,t}\in\mathbb{R}^n_{\geq0}\times\mathbb{R}~\big|~\rbrac{x_1\cdots x_n}^{\frac{1}{n}}\geq t},
\end{equation}
then maximizing the ellpsoid $\mathcal{E}$ is equivalent to maximizing a new variable $t$ with the constraint $\rbrac{\diag\rbrac{L_\mathcal{E}},t}\in\mathcal{K}_{1/n}$. As for the constraints~(\ref{eq::OrthonormalityMVIE_constraints}), we denote the SOC as $\mathcal{K}_n$ and describe the Cartesian product of $m$ SOC as $\mathcal{K}_n^m$,
\begin{gather}
\mathcal{K}_n=\cBrac{\rbrac{t,x}\in\mathbb{R}\times\mathbb{R}^{1\times n-1}~\big|~t\geq\norm{x}}\label{eq:SOC},\\
\mathcal{K}_n^m=\mathcal{K}_n\times\cdots\times\mathcal{K}_n\subseteq\mathbb{R}^{m\times n}.
\end{gather}
Then the optimization (\ref{eq:NoOrthonormalityMVIE}) is formulated as
\begin{subequations}
\label{eq:MVIE_3D}
\begin{align}
\max_{t,L_\mathcal{E},b_\mathcal{E}}&{~t},\\
\mathrm{s.t.}~&~\rbrac{\diag\rbrac{L_\mathcal{E}},t}\in\mathcal{K}_{1/n},\label{eq:MVIE_obj_constraint}\\
&~\rbrac{b_\mathcal{P}-A_\mathcal{P}b_\mathcal{E},A_\mathcal{P}L_\mathcal{E}}\in\mathcal{K}_{n+1}^m\label{eq:MVIE_SOC_constraint},\\
&~L_\mathcal{E}~\mathrm{is~lower~triangular},
\end{align}
\end{subequations}
where $m$ denotes the number of halfspaces composing $\mathcal{P}$ defined in~(\ref{eq:convex_polytope}).
Additionally, $\mathcal{K}_{1/n}$ can also be represented by SOC using additional $O(n)$ variables and cones of $\mathcal{K}_3$~\cite{Ben2001LecturesMCO}.
Eventually, we transform (\ref{eq:MVIE_3D}) into an optimization with only SOC form constraints and a simple linear objective function.

\subsection{Affine Scaling for Solving SOCP with Massive Constraints}

As we reformulate MVIE into a pure SOCP from (\ref{eq:MVIE_3D}), for brevity, we denote it as
\begin{subequations}
    \label{eq:SOCPMVIE}
    \begin{align}
    \min_{x}~&{c_\mathcal{K}\tp x}\label{eq:SOCPMVIE_obj},\\
    \mathrm{s.t.}~&({c}_i\tp x+{d}_i,~x\tp {A}_i)\in\mathcal{K}_{n_i},~1 \leq i \leq \bar{m} \label{eq:SOCPMVIE_constraint},
    \end{align}
\end{subequations}
where $x\in\mathbb{R}^{\bar{n}}$, $c_\mathcal{K}\in\mathbb{R}^{\bar{n}}$, ${c}_i\in\mathbb{R}^{\bar{n}}$, ${d}_i\in\mathbb{R}$ and ${A}_i$ are all constant.
The new decision variable $x$ consists of all lower triangular elements of $L_\mathcal{E}$, $b_\mathcal{E}$, $t$ of (\ref{eq:MVIE_3D}), and the elements added to deal with constraint (\ref{eq:MVIE_obj_constraint}) in form of hypograph of geometric mean.
We have $\bar{n}=O(n^2)$ because an $n$ dimensional ellipsoid already has $n(n+3)/2$ variables.
For convenience, we set $t$ as the $\bar{n}$-th element of $x$, 
then we have
\begin{equation}
    \label{eq:c_k}
    c_\mathcal{K}=\left(0,0,...,0,-1\right)\tp \in\mathbb{R}^{\bar{n}}.
\end{equation}
$\bar{m}=O(m+n)$ indicates the amount of the SOC constraints.
For the constraints in (\ref{eq:SOCPMVIE_constraint}) corresponding to the constraints of (\ref{eq:MVIE_obj_constraint}) which are represented by additional cones of $\mathcal{K}_3$, ${A}_i\in\mathbb{R}^{\bar{n}\times2}$ and $n_i=3$.
For the constraints in (\ref{eq:SOCPMVIE_constraint}) corresponding to the origin constraints of (\ref{eq:MVIE_SOC_constraint}), ${A}_i\in\mathbb{R}^{\bar{n}\times n}$ and $n_i=n+1$.

Then we aim to efficiently solve the SOCP~(\ref{eq:SOCPMVIE}). 
Although the dimension $n$ of the original problem~(\ref{eq:AbstractMVIE})  is limited to $2$ or $3$, the large constraint number $m$ results in a huge number of constraints $\bar{m}$ in the newly constructed SOCP. 
To tackle this multi-constraint SOCP, we generalize Affine Scaling (AS)~\cite{lagarias1990ii}, an interior-point method originally used for LP, to SOCP. 
This algorithm has an closed-form update step of each iteration and exhibits  superlinear convergence~\cite{tsuchiya1996superlinear}.
At each iteration, AS uses the logarithmic barrier function of the constraints to compute a strictly feasible region and calculates the optimal update step  within this region.
For (\ref{eq:SOCPMVIE}), its logarithmic barrier function $\phi(x)$ is defined as
\begin{gather}
    \phi(x) = -\sum_{i=1}^{\bar{m}}\log\left(f_i(x)\right),\\
    f_i(x)=\left(c_i\tp x + d_i\right)^2 - x\tp A_i A_i\tp x.
\end{gather}
Then the Hessian of $\phi(x)$ is given by
\begin{equation}
    \label{eq:SOCPDikin_Hessian}
    H_{\phi}(x) =\sum_{i=1}^{\bar{m}}\frac{1}{f_i(x)^2}\nabla f_i(x)\nabla f_i(x)\tp - \sum_{i=1}^{\bar{m}}\frac{1}{f_i(x)}\nabla^2 f_i(x),
\end{equation}
where the gradient and Hessian of $f_i(x)$ are
\begin{gather}
    \nabla f_i(x) = 2\left(c_i\tp x + d_i\right)c_i - 2A_i A_i\tp x,\\
    \nabla^2 f_i(x) = 2\left(c_i c_i\tp - A_i A_i\tp\right).
\end{gather}
Now we obtain the strictly feasible region of AS at $(k+1)$-th iteration based on the feasible $k$-th solution $x_k$ as
\begin{equation}
    \label{eq:Dikin}
    \left\{x\in\mathbb{R}^n~|~(x-x_k)\tp H_{\mathcal{\phi}}(x_k) (y-x_k) \leq 1 \right\},
\end{equation}
which is an ellipsoidal region as $H_{\mathcal{\phi}}(x_k)$ is positive definite.
Then the update step of (\ref{eq:SOCPMVIE}) can be given by
\begin{equation}
    \label{eq:Dikin_update}
    x_{k+1} = x_{k} - \tau \frac{H_{\phi}^{-1}c_{\mathcal{K}}}{\sqrt{c_{\mathcal{K}}\tp H_{\phi}^{-1}c_{\mathcal{K}}}},
\end{equation}
where $\tau\in(0,1]$ is the step size. 

\section{Solving 2-D MVIE Analytically via A Linear-time Complexity Algorithm}
\label{sec:exact_MVIE}

Inspired by the linear-time algorithm of the dual problem Minimum Volume Enclosing Ellipsoid (MVEE)~\cite{welzl2005smallest}, in this section, we focus on the construction of a linear-time algorithm for 2-D MVIE, leveraging its special LP-type problem~\cite{sharir1992combinatorial} structure. 
We begin by providing the necessary background knowledge related to LP-type problems and analyzing the limitations of existing general solutions for applying to MVIE in Sec.~\ref{sec:lp type problem}. 
Then we address these limitations and propose our algorithm with linear time complexity in Sec.~\ref{sec:Randomized Maximal Inscribed Ellipse Algorithm}.
Finally we provide the analytic solution of the subproblems required in the proposed algorithm in Sec.~\ref{sec:max_nt_ellipse}.

\subsection{Background of LP-type Problem}
\label{sec:lp type problem}
Let us consider an abstract minimization~\cite{sharir1992combinatorial} specified by pairs $(\mathcal{H}, w)$, where
$\mathcal{H}$ is a finite set of constraints and $w:2^{\mathcal{H}}\rightarrow \mathcal{W}$ is a value function that maps subsets of $\mathcal{H}$ to values in a ordered set $(\mathcal{W}, <)$, which has a unique minimum value $-\infty$.
For the sake of simplicity in subsequent descriptions, we define 
finite sets $\mathcal{G}$, $\mathcal{F}$ and a constraint $h$, which satisfy $\mathcal{G} \subseteq  \mathcal{F} \subseteq  \mathcal{H}$, $h\in \mathcal{H}$.
The problem with $(\mathcal{H}, w)$ can be considered as an LP-type problem as long as the following two properties are satisfied~\cite{sharir1992combinatorial}:
for any $\mathcal{G},\mathcal{F}$ and $h$ we have
\begin{itemize}
    \item \textit{Monotonicity}:~$w(\mathcal{G}) \leq w(\mathcal{F})$,
    \item \textit{Locality}:~with $-\infty<w(\mathcal{G})=w(\mathcal{F})$, if $w(\mathcal{F})<w(\mathcal{F}\cup\{h\})$, then $w(\mathcal{G})<w(\mathcal{G}\cup\{h\})$.
\end{itemize}
Three important definitions of LP-type problem are given:
\begin{itemize}
    \item $w(\mathcal{H})$ is called the \textit{value} of $\mathcal{H}$,
    \item constraint $h$ is \textit{violated} by $\mathcal{H}$, if $w(\mathcal{H}) < w(\mathcal{H}\cup \{h\})$,
    \item the \textit{basis} of $\mathcal{H}$ is the minimal subset of $\mathcal{H}$ with the same value of $\mathcal{H}$.
\end{itemize}
Then two primitive operations is defined:
given a basis $\mathcal{B}$,
\begin{itemize}
    \item \textit{Violation test}:~determine whether $w(\mathcal{B}) < w(\mathcal{B}\cup \{h\})$, for a constraint $h\notin \mathcal{B}$,
    \item \textit{Basis computation}:~compute the basis of $\mathcal{B}\cup\{h\}$ when $h$ is violated by the basis $\mathcal{B}$.
\end{itemize}

To give readers an intuitive understanding, we provide a specific example: for the dual problem MVEE~\cite{welzl2005smallest}, $\mathcal{H}$ is the input points set and $w(\mathcal{H})$ is the area of the minimum ellipse that can contain all the input points.

The goal of LP-type problem is to compute the basis of the input set $\mathcal{H}$ and its corresponding value $w(\mathcal{H})$.
A generalized algorithm for the LP-type problem is given by Matoušek et al.~\cite{matouvsek1992subexponential}. 
The algorithm operates by randomly selecting a constraint $h$ from the input set $\mathcal{H}$ and a known basis $\mathcal{B}\subseteq\mathcal{H}$ and performing a violation test. 
If a violation is detected, a basis of $\mathcal{B}\cup\{h\}$ is computed, then the algorithm is recursively called with the new basis and the set of checked constraints.  
This algorithm provides an efficient solution to LP-type problems with finite primitive operations, whose expected number is linear in the input set number $|\mathcal{H}|$, thanks to its randomized recursive structure.
We refer the reader interested in this complexity conclusion to Matoušek's work~\cite{matouvsek1992subexponential}.

However, for 2-D MVIE, the aforementioned generalized algorithm described above is not available for two reasons:
\textbf{i)}~In MVIE, the subset of input constraints may not form a closed polygon, which leads to an undefined ellipse.
Using the volume of the ellipse, similar to its dual problem MVEE, to define the value function $w(\cdot)$ is not feasible. 
\textbf{ii)}~There is currently no known method for performing analytical basis computation directly for any subsets in MVIE.

\begin{figure*}[t]
	\centering
	\includegraphics[width=0.95\linewidth]{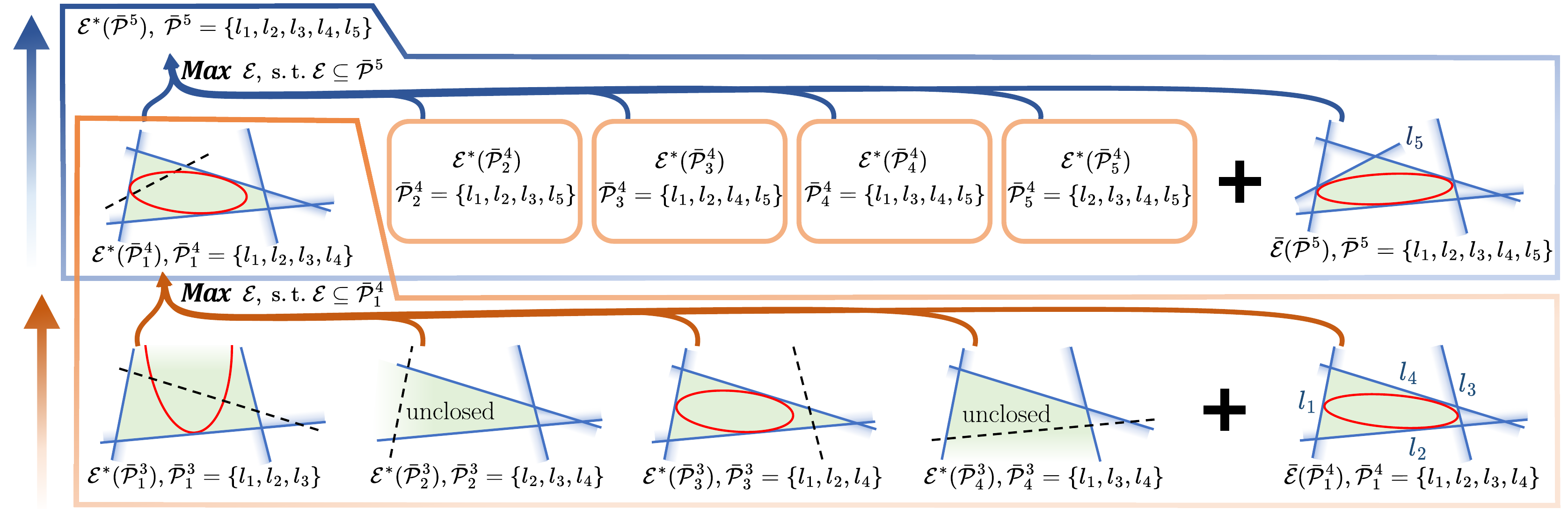}
	\vspace{-0.2cm}
	\caption{
		Illustration of a specific instance of the bottom-up strategy. 
        The large orange box specifies the process described by (\ref{eq:newMVIE example}) for calculating the MVIE of $\bar{\mathcal{P}}^4_1$ based on the MVIE of its subsets $\bar{\mathcal{P}}^3_j,1\leq j \leq 4$ and the MENN of itself.
        The red ellipses indicate MVIE $\mathcal{E}^*$ or MENN $\bar{\mathcal{E}}$, depending on the equation beneath each subset.
        Using the subset in the lower left corner as an example, the solid blue lines indicate elements in the subset $\bar{\mathcal{P}}^3_1$, the black dashed lines indicate elements in $\bar{\mathcal{P}}^4_1$ that are not included in the subset $\bar{\mathcal{P}}^3_1$.
        Similarly, the blue box illustrates the process of calculating the MVIE of $\bar{\mathcal{P}}^5$.
	}
	\label{fig:bottomup}
	\vspace{-0.4cm}
\end{figure*}

\subsection{Randomized Maximal Inscribed Ellipse Algorithm}
\label{sec:Randomized Maximal Inscribed Ellipse Algorithm}

In this subsection, we address the aforementioned challenges, and then we present a linear-time complexity algorithm by improving the generalized algorithm~\cite{matouvsek1992subexponential} for 2-D MVIE.

\subsubsection{Value Function}
\label{sec:Value Function}

We design an additional evaluation of the proximity to closure for unclosed case and use the MVIE volume for closed case to construct a new value function that satisfies the properties of the LP-type problem.
We define the value domain $\mathcal{W}$ in lexicographical ordering set $(\mathcal{W}, <)$.
We denote $<$ as the lexicographical ordering on $\mathbb{R}^2$, that is, for $\forall x=(x_1,x_2)\tp,y=(y_1,y_2)\tp\in \mathbb{R}^2$, $x < y$, if $x_1 < y_1$, or $x_1 = y_1$ and $x_2 < y_2$.
This ordering is extended to $\mathbb{R}^2\cup \{-\infty\}$ by the definition that $-\infty$ is the unique minimum value.
Considering the frequent failure to form closed polygon for the subsets of constraints, we define the value function as
\begin{equation}
    \label{eq:value_function}
    w(\mathcal{G})=
    \left\{
        \begin{aligned}
            &-\infty&  &\mathit{if}~\mathcal{G}=\varnothing  \\
            &\bar{w}_u(\mathcal{G})& &\mathit{if}~\mathcal{G}~\mathit{is~unclosed} \\
            &\bar{w}_c(\mathcal{G})& &\mathit{if}~\mathcal{G}~\mathit{is~closed}
        \end{aligned}
    \right.,
\end{equation}
where $\bar{w}_u(\mathcal{G})$ represents a measure of the proximity to closure when $\mathcal{G}$ is not closed.
To define $\bar{w}_u(\mathcal{G})$, we first define a new set $\mathcal{G}_n$ that does not contain duplicate elements, and all elements in the new set $\mathcal{G}_n$ come from the unit normal vectors corresponding to the halfspace constrain elements in the set $\mathcal{G}$.
Then we define $\bar{w}_a(\mathcal{G})$ in lexicographical ordering by
\begin{equation}
    \label{eq:value_function:unclosed}
    \bar{w}_u(\mathcal{G})=(-a_p(\mathcal{G}_n),~\min\{|\mathcal{G}_n|,3\}),
\end{equation}
where $a_p(\mathcal{G}_n)$ denotes the angle of polar cone of $\mathcal{G}_n$.
Such a particular definition is intended to deal with the corner cases that arise due to the parallel elements in $\mathcal{G}$.

As for $\bar{w}_c(\mathcal{G})$ when $\mathcal{G}$ is closed, we define it by
\begin{equation}
    \bar{w}_c(\mathcal{G})=(\area(\mathcal{E}^*(\mathcal{G}))^{-1},~\xi ),
\end{equation}
whose first element represents the inverse of the MVIE's area of $\mathcal{G}$, and the second element $\xi$ serves to extend this function to $\mathbb{R}^2$.
The first element of $\bar{w}_c(\mathcal{G})$ is always greater than 0, and the first element of $\bar{w}_u(\mathcal{G})$ is always less than or equal to 0. 
Thus no matter what value $\xi$ takes, we have $\bar{w}_c > \bar{w}_u$.
Now based on this value function that satisfies monotonicity and locality, whose proof is detailed in Appendix~\ref{apd:Proof of Monotonicity} and \ref{apd:Proof of Locality}, MVIE can be treated as an LP-type problem.

\subsubsection{Basis Computation}
\label{sec:Basis Computation}
To perform analytical basis computation for MVIE, we decompose MVIE into several subproblems that can be solved analytically, then MVIE of $\mathcal{P}$ defined in (\ref{eq:AbstractMVIE}) is equivalent to
\begin{equation}
    \label{eq:newMVIE}
    \mathcal{E}^{*}(\mathcal{P}) = \max_{{\bar{\mathcal{P}}^N}\subseteq\mathcal{P}}{\area \left({\bar{\mathcal{E}}(\bar{\mathcal{P}}^N)}\right)},~\mathrm{s.t.}~\bar{\mathcal{E}}\subseteq\mathcal{P},
\end{equation}
where $\mathcal{P}$ is the input polygon \rv{ while simultaneously denoting the finite set of the halfspaces whose intersection constitutes the input polygon in this section. 
Its subset containing $N$ elements is denoted as ${\bar{\mathcal{P}}^N}$.
To prevent potential misunderstanding, it should be emphasized that the subset ${\bar{\mathcal{P}}^N}$ is derived from $\mathcal{P}$, the sole input of MVIE, and is independent of the obstacles used to generate $\mathcal{P}$ in RsI.}
Additionally, $\bar{\mathcal{E}}(\bar{\mathcal{P}}^N)$ is the subproblem defined by
\begin{subequations}
    \label{eq:subMVIE}
    \begin{align}
        \bar{\mathcal{E}}(\bar{\mathcal{P}}^N) = \max_{\mathcal{E}}~&{\area \rbrac{\mathcal{E}}},\\
    \mathrm{s.t.}~~&\mathcal{E}\subseteq{\bar{\mathcal{P}}^N},\\
    &\mathcal{E}~\mathrm{is~tangent~to}~h,~\forall~h\in{\bar{\mathcal{P}}^N}\label{eq:subMVIE:constraint}.
    \end{align}
\end{subequations}
Since at least $3$ halfspaces are required to form a closed space in 2-D and an ellipse has only $5$ degrees of freedom, we require $3\leq N\leq 5$.
When ${\bar{\mathcal{P}}^N}$ is closed and has no redundant element, (\ref{eq:subMVIE}) has an analytical solution, and its result $\bar{\mathcal{E}}$ is the \textbf{M}aximal \textbf{E}llipse tangent to \textit{\textbf{N}} edges of the \textit{\textbf{N}}-gon formed by ${\bar{\mathcal{P}}^N}$, denoted as \textbf{MENN}, detailed in Sec.~\ref{sec:max_nt_ellipse}.
Notably, $\mathcal{E}^*$ and $\bar{\mathcal{E}}$ are used to denote MVIE and MENN, respectively.
For the two cases where subproblems~(\ref{eq:subMVIE}) are infeasible, we handle them with special treatment:
\textbf{i)}~If ${\bar{\mathcal{P}}^N}$ is unclosed, we define its $\bar{\mathcal{E}}$ as $\infty$.
\textbf{ii)}~If there are redundant elements in a closed ${\bar{\mathcal{P}}^N}$, that is, these elements are not any side of the formed closed polygon.
At this point, it is impossible for these elements to satisfy (\ref{eq:subMVIE:constraint}), then we also define its $\bar{\mathcal{E}}$ as  $\infty$.
Based on the subproblem~(\ref{eq:subMVIE}), MVIE~(\ref{eq:newMVIE}) can be solved through enumerating subproblems~(\ref{eq:subMVIE}) which has analytical solution.

Inspired by the distance algorithm of GJK~\cite{montanari2017improving}, we organize the enumeration in an orderly way by using a bottom-up strategy, about which we illustrate a detailed example in Fig.~\ref{fig:bottomup}.
In Fig.~\ref{fig:bottomup}, we use subscript to distinguish between different subsets of a set, for instance, ${\bar{\mathcal{P}}^4_1}$ denotes one of the subsets of ${\bar{\mathcal{P}}^5}$.
Specifically regarding the bottom-up strategy, for a subset ${\bar{\mathcal{P}}^4_1}$ containing $4$ elements, we compute its MVIE by
\begin{subequations}
    \label{eq:newMVIE example}
    \begin{align}
    \mathcal{E}^{*}(\bar{\mathcal{P}}^4_1) = \max_{\mathcal{E}}~&{\area \rbrac{{\mathcal{E}}}},\\
    \mathrm{s.t.}~~&{\mathcal{E}}\subseteq \bar{\mathcal{P}}^4_1,\label{eq:newMVIE example cons}\\
    &{\mathcal{E}} \in \{\bar{\mathcal{E}}(\bar{\mathcal{P}}^4_1)\}\cup\{\mathcal{E}^{*}(\bar{\mathcal{P}}^3_j)\}, 1\leq j \leq 4,
\end{align}
\end{subequations}
where $\bar{\mathcal{P}}^3_j$ denotes a subset, which contains $3$ elements, of $\bar{\mathcal{P}}^4_1$.
\rv{Notably, the MVIE of a closed $\bar{\mathcal{P}}^3_j$ is exactly  its MENN~\cite{minda2008triangles}.}
This equation~(\ref{eq:newMVIE example}) corresponds to the process in the large orange box in Fig.~\ref{fig:bottomup}:
among the MVIE for each subset $\bar{\mathcal{P}}_j^3$, and the MENN of $\bar{\mathcal{P}}^4_1$, we select the largest one that satisfies the constraint~(\ref{eq:newMVIE example cons}) as the MVIE of $\bar{\mathcal{P}}_1^4$. 
Based on the result we can build upwards MVIE of $\mathcal{P}^5$ containing $5$ elements in a similar way, as illustrated in the blue box of Fig.~\ref{fig:bottomup}.
\rv{It should be emphasized that for the actual input $\mathcal{P}$, which typically consists of more than $5$ halfspaces, its MVIE is also computed through a similar bottom-up enumeration: calculating the MVIE of all its subsets $\bar{\mathcal{P}}^5$ and selecting the largest ellipse contained within $\mathcal{P}$.
To summarize, by connecting the above processes through the bottom-up strategy, now we can compute the MVIE of the input $\mathcal{P}$ based entirely on its subsets' MENN ($3\leq N\leq 5$) which can be solved analytically in an efficient enumeration order.}

\begin{algorithm}[t]
    \label{alg:MaxEllpise}
    \DontPrintSemicolon
    \LinesNumbered
    \SetAlgoLined
    \caption{\textbf{MaxEllipse}}
    \textbf{Notion:} combinatorial dimension $\delta=5$\;
    \KwIn{~~$\mathcal{H}$: input set,~~$\mathcal{X}$: a subset of the basis of $\mathcal{H}$}
    \KwOut{$\mathcal{B}$: basis of $\mathcal{H}$,~~$v_\mathcal{B}$: value of $\mathcal{B}$}
    \Begin
    {
        $v_\mathcal{X}\leftarrow w(\mathcal{X})\label{alg:MIEUpdateStart}$~~
        $\textcolor{blue}{\text{/* defined in (\ref{eq:value_function}) */}}$\;
        \If{$|\mathcal{X}| \geq \delta \label{alg:combination dimension}$} 
        {
            \Return{$\mathcal{X},v_\mathcal{X}$}
        }
        $\mathcal{S}\leftarrow \{\}$\;
        $\mathcal{B}\leftarrow \mathcal{X}$\;
        $v_\mathcal{B} \leftarrow v_\mathcal{X} $\;
        \ForEach{$h\in\mathcal{H}~~\mathrm{in~a~random~order}\label{alg:MIE_loop_start}$}
        {
            \If{$\mathbf{ViolateTest}(h,v_\mathcal{B}, \mathcal{X})\label{alg:MIE_Violate}$}
            {
                $\mathcal{B},v_\mathcal{B} \leftarrow \textbf{MaxEllipse}(\mathcal{S},\mathcal{X}\cup\{h\})\label{alg:MIE_MIE}$\;
            }
            $\mathcal{S}\leftarrow \mathcal{S}\cup\{h\}\label{alg:MIE_loop_end}$\;
        }
        \Return{$\mathcal{B},v_\mathcal{B}$}
    }
\end{algorithm}

\subsubsection{Algorithm Overview}
Building upon the aforementioned efforts, we improve the general framework~\cite{matouvsek1992subexponential} to propose a randomized algorithm for solving 2-D MVIE as shown in Algorithm~\ref{alg:MaxEllpise}.
For the initial call to Algorithm~\ref{alg:MaxEllpise}, we set $\mathcal{H}\leftarrow \mathcal{P}$ and $\mathcal{X}\leftarrow \{\}$.
The recursive framework (Line~\ref{alg:MIE_MIE}) of the algorithm, combined with the previously mentioned bottom-up strategy, results in an effect:
when a recursive call occurs, it indicates that all subsets of the input $\mathcal{X}$ have already been checked, and for any of its subsets $\bar{\mathcal{X}}$, there is always an element $h\in\mathcal{X}$ violated by $\bar{\mathcal{X}}$. 
Therefore, if the input $\mathcal{X}$ is closed, we can directly solve the MENN of the polygon formed by $\mathcal{X}$ for the value $w(\mathcal{X})$ in Line~\ref{alg:MIEUpdateStart}. 
And if $\mathcal{X}$ is unclosed, it is easy to use (\ref{eq:value_function:unclosed}) to compute its value.
Recalling the complexity conclusion we mentioned in Sec.~\ref{sec:lp type problem}, now we can obtain the result of Algorithm~\ref{alg:MaxEllpise} by finite number of violation tests (Line~\ref{alg:MIE_Violate}), MENN (for closed case) and (\ref{eq:value_function:unclosed}) (for unclosed case), whose expected number is linear in the input set number.

In addition, the maximum cardinality of any basis is denoted as combinatorial dimension $\delta$, and based on the value function (\ref{eq:value_function}) we have $\delta=5$.
This implies that in Line~\ref{alg:MIEUpdateStart}, we will only encounter MENN calculations for triangles, quadrilaterals, and pentagons, which also consistent with our decomposition of MVIE in (\ref{eq:subMVIE}).
The combinatorial dimension $\delta$ in Line~\ref{alg:combination dimension} ensures that the maximum number of recursive iterations in the algorithm remains a constant.

The analytical solution method (detailed in Sec.~\ref{sec:max_nt_ellipse}) may lead to incorrect result when the input is redundant.
Thus, as mentioned in Sec.~\ref{sec:Basis Computation}, by treating redundant cases separately and assigning them a solution of $\infty$, we prevent them from influencing the overall algorithm.
In Algorithm~\ref{alg:MaxEllpise}, we add this feature in ViolateTest (Line~\ref{alg:MIE_Violate}):
in addition to the evaluation involved $h$ and $v_\mathcal{B}$ based on (\ref{eq:value_function}), in the case where $h$ is violated by $\mathcal{B}$, if $\mathcal{X}\cup \{h\}$ forms a convex $N$-gon but $N < |\mathcal{X}\cup \{h\}|$ (redundant case), then false is returned.
Since we perform checks based on the bottom-up strategy, the non-redundant closed subsets (if they exist) of the redundant closed set $\mathcal{X}\cup \{h\}$ will definitely be checked in the algorithm.
Thus there is no need to check with the redundant closed set.

Now we construct the linear complexity algorithm for 2-D MVIE.
Another factor that affects the practical efficiency of the algorithm is the solution of MENN~(\ref{eq:subMVIE}), for which we present efficient analytical solutions in Sec.~\ref{sec:max_nt_ellipse}.

\subsection{Maximal Ellpise tangent to N edges of the N-gon}
\label{sec:max_nt_ellipse}
As demanded in Algorithm~\ref{alg:MaxEllpise}, in this section we present the analytical computation of MENN, which is the maximal inscribed ellipse that is tangent to all edges of arbitrary convex $N$-gon, $N\in\{3,4,5\}$.
For arbitrary non-degenerate triangle~\cite{minda2008triangles}, convex quadrilateral~\cite{horwitz2005ellipses} or convex pentagon~\cite{agarwal2015duality}, there exists a unique such ellipse.
Since the inputs from Algorithm~\ref{alg:MaxEllpise} are non-degenerate and convex, in the following we default to the existence and uniqueness of the MENN.

Referring to the definition in (\ref{eq:ellipsoid}), in the 2-D case we define the point $p\in\mathbb{R}^2$ on the boundary of the ellipse $\mathcal{E}$ to satisfy
\begin{equation}
    \label{eq:ellipse}
    (p-b_\mathcal{E})\tp (A_\mathcal{E}D_\mathcal{E}^2A_\mathcal{E}\tp)^{-1} (p-b_\mathcal{E}) = 1,
\end{equation}
We denote the \textit{homogeneous coordinate}~\cite{jones1912introduction} of point $p$ as $\hat{p}=\left(p\tp,~1\right)\tp$.
Since $A_\mathcal{E}D_\mathcal{E}^2A_\mathcal{E}\tp$ is positive definite symmetric, (\ref{eq:ellipse}) is a second-degree polynomial equation in two elements of $p$.
We transform (\ref{eq:ellipse}) into a quadratic form, then the ellipse $\mathcal{E}$ can be described in homogeneous coordinate as
\begin{equation}
    \label{eq:ellipse_homogeneous}
    \mathbb{P}=\cBrac{\hat{p}\in\mathbb{R}^3 ~\big| ~\hat{p}\tp M_P \hat{p} = 0},
\end{equation}
where the coefficient matrix $M_P\in\mathbb{R}^{3\times3}$ is symmetric. 

For the MENN problem in this section, it is intractable to solve the problem by (\ref{eq:ellipse_homogeneous}) which demand the points of tangency. 
Thus we adopte to leverage the information of the tangents directly, based on the polarity of points and lines with respect to the ellipse $\mathcal{E}$~\cite{richter2011conics}.
First, we require an algebraic characterization of a line.
Specifically, in the projective plane, given a point with a homogeneous coordinate  $\hat{p}$, the line passing through the point $\hat{p}$ can be denote in the form of \textit{line corrdinate}~\cite{jones1912introduction} as $l\in\mathbb{R}^{3}$ that satisfies
\begin{equation}
\label{eq:PL_relationship}
    \hat{p}\tp l=0,
\end{equation}
based on which, the calculation of the line coordinate can be performed by the Grassmanian expansion~\cite{klein1939elementary}.
According to the polarity of ellipse~\cite{richter2011conics}, when $\hat{p}$ is on the ellipse $\mathcal{E}$ (\ref{eq:ellipse_homogeneous}), the line coordinate $l$ of the line tangent to $\mathcal{E}$ at $\hat{p}$ is given by
\begin{equation}
	\label{eq:tanget_line}
	l=M_P \hat{p}.
\end{equation}
Since the ellipse is not degenerate, $M_P$ is invertible.
Based on (\ref{eq:ellipse_homogeneous},~\ref{eq:tanget_line}), similarly to the representation by a set of the points in (\ref{eq:ellipse_homogeneous}), the ellipse $\mathcal{E}$ now can be described by the set of all its tangents in line coordinate as
\begin{gather}
    \mathbb{L}=\cBrac{l\in\mathbb{R}^3 ~\big| ~l\tp M_L l = 0}\label{eq:ellipse_homogeneous_line},\\
    M_L \propto {M_P}^{-1} \label{eq:ellipse_conversion}.
\end{gather}

Now we can calculate the MENN by utilizing the information of the tangents directly.
Specifically, we first compute the line coordinates of edges of the polygon for solving $M_L$, then obtain $M_P$, and eventually get the ellipse in the desired form of (\ref{eq:ellipsoid}).
This process involves using (\ref{eq:PL_relationship}), (\ref{eq:ellipse_homogeneous_line}), (\ref{eq:ellipse_conversion}), (\ref{eq:ellipse}) and (\ref{eq:ellipse_homogeneous}) sequentially.
In the following, we present the implementation details for different $N$.

\subsubsection{MENN of a Convex Pentagon}
\label{sec:Convex_Pentagon}
For (\ref{eq:ellipse_homogeneous_line}), the symmetric $M_L$ has $6$ variables.
Bringing the line coordinates of the five edges of the convex pentagon into (\ref{eq:ellipse_homogeneous_line}) respectively, we can obtain a six-element homogeneous system of linear equations consisting of five equations. 
Since there exists and only exists unique ellipse tangent to all edges of the convex pentagon~\cite{agarwal2015duality}, the homogeneous system always has a non-trivial solution.
Then based on the solution,  we can obtain the MENN progressively through the process aforementioned.

\subsubsection{MENN of a Convex Quadrilateral}
\label{sec:Convex_Quadrilateral}
\begin{figure}[!t]
	\centering
	\includegraphics[width=1\linewidth]{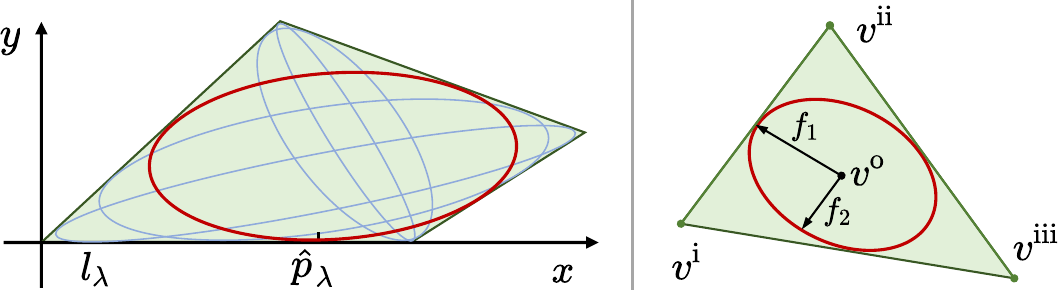}
	\vspace{-0.6cm}
    \caption{
		\textbf{Left}: Illustration of the calculation of the MENN of a convex quadrilateral.
        The blue and red ellipses represent the ellipses obtained by sampling different points $\hat{p}_\lambda$ on the coinciding edge $l_\lambda$.
        \textbf{Right}: Illustration of the calculation of the MENN of a triangle.
	}
	\label{fig:quadrilateral}
	\vspace{0.2cm}
\end{figure}

In contrast to convex pentagon, a convex quadrilateral only provides four tangents.
Thus there are a unique one-parameter set of inscribed ellipses that are tangent to all edges of the quadrilateral, whose parameter can be taken to be a prescribed point contact on any single edge of the quadrilateral~\cite{agarwal2015duality} as shown in Fig.~\ref{fig:quadrilateral}.
Additionally, only one of them has the largest area~\cite{horwitz2005ellipses}, which is the MENN of the quadrilateral.

For simplicity of calculation, as shown in Fig.~\ref{fig:quadrilateral}, we translate the quadrilateral so that one of its vertices coincides with the origin and one of the edges connected to that vertex coincides with the $x$-axis, which will not change the shape of the quadrilateral.
Then we denote the line coordinate of the coinciding edge as $l_\lambda$.
Inspired by the work of Hayes~\cite{hayes2019largest}, we introduce new constraint by using the point of tangency on $l_\lambda$, and denote the point as $\hat{p}_\lambda$.
Then we have
\begin{equation}
	\hat{p}_\lambda=(\lambda, 0, 1)\tp,~~l_\lambda=(0,1,0)\tp,
\end{equation}
where $\lambda$ is a variable.
Combining (\ref{eq:tanget_line}) and (\ref{eq:ellipse_conversion}), their polarity relationship can be written as 
\begin{equation}
    M_L l_\lambda \propto \hat{p}_\lambda,
\end{equation}
which is an independent new constraint on $M_L$.
The subsequent operations are similar to Sec.~\ref{sec:Convex_Pentagon}, with the difference that eventually we calculate the area~\cite{hayes2019largest} of $\mathcal{E}$ as a function $A_\mathcal{E}(\lambda)$ in terms of $\lambda$.
The optimal $\lambda$ corresponding to the MENN of the quadrilateral can be computed by calculating the zeros of the first order derivative of $A_\mathcal{E}(\lambda)$. 
This calculation only involves solving a quadratic equation w.r.t. $\lambda$, which can be solved quickly and analytically.
Due to space limitations, we do not delve into further details of the calculation.

\subsubsection{MENN of a Triangle}
\label{sec:Triangle}
The Steiner inellipse tangent to the edges of the triangle at their midpoints, is the MENN~\cite{minda2008triangles}.
As shown in Fig.~\ref{fig:quadrilateral}, we denote the 2-D Cartesian coordinates of the vertices of the triangle as $v^{*}=(v^{*}_0, v^*_1)\tp, *=\{\textrm{i},\textrm{ii},\textrm{iii}\}$.
The center $v^o$ and two conjugate diameters $f_1,f_2\in\mathbb{R}^2$ of the Steiner inellipse can be written as
\begin{gather}
    v^o = \frac{1}{3}(v^\textrm{i}+v^\textrm{ii}+v^\textrm{iii}),\\
    f_1 = \frac{1}{2}(v^o-v^\textrm{iii}),~f_2 = \frac{1}{2 \sqrt{3}}(v^\textrm{i}-v^\textrm{ii}).
\end{gather}
Then the parameters required in (\ref{eq:ellipsoid}) can be calculated by
\begin{equation}
    b_\mathcal{E}=v^o,~A_\mathcal{E}D_\mathcal{E}^2A_\mathcal{E}\tp=
    \bigl(f_1~ f_2\bigr)\bigl(f_1~ f_2\bigr)\tp.
\end{equation}

\section{Evaluation and Benchmark}

To comprehensively demonstrate the outstanding advantages of FIRI, in terms of efficiency, quality, and manageability, we design extensive benchmarks comparing FIRI against several state-of-the-art convex polytope generation algorithms.
Additionally, the efficiency of FIRI lies in the methods that we develop specifically for the two optimizations (Line~\ref{algl:RestrictiveInflationStart} and~\ref{algl:EllipsoidProgramming} of Algorithm~\ref{alg:FIRI}).
To demonstrate their effectiveness, we conduct comparative experiments for these methods which are proposed for solving minimum-norm with small dimension but massive constraints and for solving MVIE, respectively.
\rv{All benchmarks run on Intel Core i7-12700KF CPU, Linux, using C++14 without hardware acceleration.}

\begin{table}[t]
    \renewcommand\arraystretch{1.2}
    \tabcolsep=0.2cm
    \caption{\rv{Comparison of Adaptability of Different Methods for Generating Free Convex Polytope}}
    \label{tab:input capablilty}
    \centering
    \begin{tabular}{c c c c c c }
        \toprule
        \multirow{2}{*}{\diagbox[width=2.7cm]{Method}{Adaptability}} & \multicolumn{3}{c}{Seed Type} & \multicolumn{2}{c}{Obstacles Type}\\
		\cmidrule(lr){2-4} \cmidrule(lr){5-6} 
        & ~~Point~~ & Line &  Polytope & Point &  Polytope  \\
        \toprule
        \textbf{FIRI} & \color{teal}\CheckmarkBold & \color{teal}\CheckmarkBold & \color{teal}\CheckmarkBold & \color{teal}\CheckmarkBold & \color{teal}\CheckmarkBold \\

        IRIS\cite{Deits2015ComputingIRIS} & \color{teal}\CheckmarkBold & \color{red!75}\XSolidBrush & \color{red!75}\XSolidBrush & \color{teal}\CheckmarkBold & \color{teal}\CheckmarkBold \\

        Galaxy\cite{Zhong2020GeneratingLCP} & \color{teal}\CheckmarkBold & \color{red!75}\XSolidBrush & \color{red!75}\XSolidBrush & \color{teal}\CheckmarkBold & \color{red!75}\XSolidBrush \\

        RILS\cite{Liu2017PlanningDF} & \color{teal}\CheckmarkBold & \color{teal}\CheckmarkBold & \color{red!75}\XSolidBrush & \color{teal}\CheckmarkBold & \color{red!75}\XSolidBrush \\
        \toprule
    \end{tabular}
\end{table}

\afterpage{
\begin{table*}[b]
    \renewcommand\arraystretch{1.2}
    \tabcolsep=0.12cm
    \caption{\rv{Success Rate of Different Methods for Generating Convex Polytopes Containing Seed across Different Scenarios and Seed Types}}
    \label{tab:success rate of manageability}
    \centering
    \begin{tabular}{c c c c c c c c c c c c c c }
        \toprule
        \multicolumn{2}{c}{\multirow{4}{*}{{Scenario}}} & \multicolumn{12}{c}{Success Rate $[\%]$} \\
        \cmidrule(lr){3-14}
        & & \multicolumn{4}{c}{Point Seed} & \multicolumn{4}{c}{Line Seed} & \multicolumn{4}{c}{Polytope Seed}  \\
		\cmidrule(lr){3-6} \cmidrule(lr){7-10} \cmidrule(lr){11-14} 
        &  & ~~~\textbf{FIRI}~~~ & IRIS\cite{Deits2015ComputingIRIS} & Galaxy\cite{Zhong2020GeneratingLCP}  & RILS\cite{Liu2017PlanningDF} & ~~~\textbf{FIRI}~~~ & IRIS\cite{Deits2015ComputingIRIS} & Galaxy\cite{Zhong2020GeneratingLCP}  & RILS\cite{Liu2017PlanningDF} & ~~~\textbf{FIRI}~~~ & IRIS\cite{Deits2015ComputingIRIS} & Galaxy\cite{Zhong2020GeneratingLCP}  & RILS\cite{Liu2017PlanningDF} \\
        \toprule
        \multirow{3}{*}{\makecell[c]{2-D}} & Sparse & \textbf{100} & 98.4 & 100 & 100 & \textbf{100} & 74.7 & 81.6 & 100 & \textbf{100} & 88.9 & 95.8 & 97.0 \\
        & Medium & \textbf{100} & 97.0 & 100 & 100 & \textbf{100} & 53.9 & 64.2 & 100 & \textbf{100} & 79.7 & 91.1 & 95.3 \\
        & Dense  & \textbf{100} & 96.6 & 100 & 100 & \textbf{100} & 48.5 & 39.1 & 100 & \textbf{100} & 69.7 & 73.9 & 90.6 \\
        \hline
        \multirow{3}{*}{\makecell[c]{3-D}} & Sparse & \textbf{100} & 99.1 & 100 & 100 & \textbf{100} & 96.5 & 79.1 & 100 & \textbf{100} &  96.2 & 85.1 & 98.0 \\
        & Medium  & \textbf{100} & 97.8 & 100 & 100 & \textbf{100} & 78.2 & 61.3 & 100 & \textbf{100} & 70.6 & 78.0 & 88.3 \\
        & Dense   & \textbf{100} & 96.2 & 100 & 100 & \textbf{100} & 59.6 & 47.0 & 100 & \textbf{100} & 37.3 & 45.1 & 70.6 \\ 
        \toprule
    \end{tabular}
\end{table*} 
}

\begin{table}[t]
    \renewcommand\arraystretch{1.2}
    \tabcolsep=0.35cm
    \caption{\rv{Input Obstacles Number of Different Scenarios}}
    \label{tab::input obstacle}
    \centering
    \begin{tabular}{c c c c c c}
        \toprule
        \multicolumn{2}{c}{\multirow{2}{*}{{Scenario}}} & \multicolumn{4}{c}{Input Obstacle Number} \\
        \cmidrule(lr){3-6}
        && avg & std & min & max \\
        \toprule
        \multirow{3}{*}{\makecell[c]{2-D}} & Sparse&246.7&111.0&44&525
        \\
        & Medium &1157.6&277.8&553&1683
        \\
        & Dense &3007.5&515.2&1443&4048
        \\
        \hline
        \multirow{3}{*}{\makecell[c]{3-D}} & Sparse & 453.6&119.6&303&670\\
        & Medium &2677.8&613.6&1524&3929\\
        & Dense &12659.0&764.9&11032&14536\\
        \toprule
    \end{tabular}
\end{table}

\subsection{Comparison of Generating Free Convex Polytope}
\label{sec:Comparison of Generating Free Convex Polytope}
Based on the description and analysis of several state-of-the-art algorithms for generating free convex polytopes in Sec.~\ref{sec::Related Work}, we benchmark the proposed FIRI against IRIS~\cite{Deits2015ComputingIRIS}, Galaxy~\cite{Zhong2020GeneratingLCP} and RILS~\cite{Liu2017PlanningDF}.
Since Gao's method~\cite{Gao2020TeachRepeatReplanAC} relies on modeling the environment as a grid map and can only operate in near real-time with low map resolution, it is not compared here.
Additionally, Galaxy can be regarded as an enhancement of Savin's method~\cite{savin2017algorithm}, both of which are based on space inversion, thus we directly adopt Galaxy for benchmark.
For IRIS and our proposed method FIRI, we set the same parameter $\rho=0.02$, that is, the stopping condition of them is that the volume of the MVIE of the convex hull obtained in this iteration grows less than $2\%$ from the last iteration.
For Galaxy and RILS, we used the default parameters.

To begin with, we compare the capability of these methods to adapt to different types of seed and obstacle inputs.
As evidenced by the results in Tab.~\ref{tab:input capablilty}, FIRI demonstrates the highest level of adaptability to various inputs.
\rv{In Tab.~II, the term of adaptability of polytope-type obstacles denotes the ability to handle polytope obstacles directly without discretization. 
As polytope representation is a widely adopted approach for environmental modeling~\cite{lin2023immesh,wang2023linear}, the capability to process polytope obstacles directly is worthwhile.}
Subsequently, based on the reported adaptability, we conduct extensive benchmarks to validate the superior performance of FIRI.

\subsubsection{Manageability}
\label{sec::manageability result}
As analyzed in Sec.~\ref{sec:intro}, manageability is crucial in many applications.
For instance, there are situations where we require the convex hull to contain the line segment of the path generated by the frontend~\cite{ji2022elastic}, or during whole body planning, we demand the convex hull to encompass the robot~\cite{han2023efficient}. 
Thus we compare the manageability of various methods by using point, line, and convex polytope as the seed.
Although the type of obstacles has no effect on manageability, for fairness, the obstacles input are characterized by points.

We conduct benchmark in a complex environment of $50\times50~m$ size (with a height of $10m$ for the 3-D case), where we generate random obstacles by using Perlin noise~\cite{hart2001perlin}.
For each test, we randomly generate a collision-free seed in the environment as input. 
The boundary of each convex hull generation algorithm is constrained to be a square (cube for 3-D) with side length $6~m$ centered on the seed's center and parallel to the coordinate axis.
And the obstacle input is the points within the boundary of the square in the map.
When the seed input is a line, we set its length to $1.5~m$. 
When the seed input is a convex polytope, we set it as a $1.5\times0.75\times0.25~m$ rectangular for the 3-D case, or a $1.5\times0.75~m$ rectangle for the 2-D case.
We generate the experimental environments with different obstacle densities, whose numbers of input obstacles are shown in Tab.~\ref{tab::input obstacle}. 
For each density, we create different $10$ environments. 
Furthermore, for each environment, we conduct $500$ random trials for each type of seed input.

To be fair, we perform several adjustments in different seed cases based on Tab.~\ref{tab:input capablilty}.
When the seed is a line, for Galaxy and IRIS, which can only use one point as the seed input, we use both the endpoints and midpoints of the line as seed inputs for them to compute $3$ convex polytopes.
We then select the convex hull with the highest degree of line containment as the result.
When the seed is a convex polytope, FIRI can directly take the polytope as input.
IRIS and Galaxy take each vertex and the center point of the seed as inputs to compute multiple corresponding convex hulls.
For RILS, which is adapted to use a line as input, we compute multiple convex hulls using the seed's diagonals as inputs.
For each of the above three methods, we choose the convex hull that maximizes the inclusion of the seed as the respective result.

We calculate the success rate in generating a convex polytope that fully contains the seed, as shown in Tab.~\ref{tab:success rate of manageability}.
In addition, for a more intuitive presentation, we present an example of the results of each method in an application of whole-body planning that has a need for manageability. 
As shown in Fig.~\ref{fig:maze2}, we use a rectangular robot as the seed, and the details of the application will be described in Sec.~\ref{sec:application car maze}.
Due to greedy approach of IRIS in seeking the largest possible volume of the convex hull, it performs poorly in terms of manageability, to the extent that it cannot guarantee to contain the seed even when the seed is just a single point.
While RILS can ensure to contain the line seed, it is not sufficient to guarantee that the generated convex hull contains the seed when the seed is represented as a convex hull.
As for Galaxy, its heuristic cut method does not ensure that the seed is included either, when the seed is a line or a convex polytope.
\rv{In contrast, although FIRI employs a similar approach to IRIS and RILS in computing halfspaces through ellipsoid inflation for polytope generation, the integration of our proposed RsI provides a distinct advantage: each halfspace computed by FIRI is guaranteed to contain the seed. 
This fundamental property ensures that the resulting polytope invariably encompasses the seed, thus maintaining manageability.}

\begin{table*}[t]
    \renewcommand\arraystretch{1.2}
    \tabcolsep=0.19cm
    \caption{\rv{Comparison of Computation Time of Different Methods for Generating Free Convex Polytope}}
    \label{tab:Computation Time of convex hull}
    \centering
    \begin{tabular}{c c c c c c c c c c c c c c c}
        \toprule
        \multirow{4}{*}{\makecell[c]{Seed\\Type}} & \multirow{4}{*}{\makecell[c]{Scenario}} & \multirow{4}{*}{Method} &\multicolumn{12}{c}{Computation Time $[ms]$}\\
        \cmidrule(lr){4-15}
        &&& \multicolumn{4}{c}{Sparse} & \multicolumn{4}{c}{Medium} & \multicolumn{4}{c}{Dense} \\
		\cmidrule(lr){4-7} \cmidrule(lr){8-11} \cmidrule(lr){12-15} 
        &&& avg & std & min & max & avg & std & min & max & avg & std & min & max \\
        \toprule
        \multirow{10}{*}{\makecell[c]{Point}} & \multirow{5}{*}{\makecell[c]{2-D}} & \textbf{FIRI} & \bf 0.038&\bf 0.013&\bf 0.013&\bf 0.069&\bf 0.120&\bf 0.035&\bf 0.032&\bf 0.189&\bf 0.273&\bf 0.068&\bf 0.127&\bf 0.451\\
        & & IRIS\cite{Deits2015ComputingIRIS}  &34.444&2.978&24.997&40.125&37.730&2.982&28.001&46.620&39.671&5.111&20.848&48.155\\
        \cdashline{3-15}
        & & Galaxy\cite{Zhong2020GeneratingLCP}  &0.069&0.013&0.042&0.099&0.123&0.023&0.077&0.168&0.233&0.038&0.143&0.327\\
        & & RILS\cite{Liu2017PlanningDF}  &0.011&0.004&0.005&0.021&0.037&0.010&0.016&0.061&0.082&\bf 0.017&0.027&0.122\\
        & & FIRI(SI) & \bf 0.008&\bf 0.003&\bf 0.002&\bf 0.017&\bf 0.032&\bf 0.010&\bf 0.014&\bf 0.052&\bf 0.066&0.021&\bf 0.016&\bf 0.103\\
        \cmidrule(lr){2-15} 
        & \multirow{5}{*}{\makecell[c]{3-D}} & \textbf{FIRI} & \bf 0.143& \bf 0.035& \bf 0.069& \bf 0.237& \bf 0.660& \bf 0.224& \bf 0.259& \bf 1.259& \bf 2.116& \bf 0.560& \bf 1.109& \bf 3.535\\
        & & IRIS\cite{Deits2015ComputingIRIS}  &34.638&7.083&16.966&76.988&55.724&15.327&11.173&103.176&87.897&20.823&14.826&167.441\\
        \cdashline{3-15}
        & & Galaxy\cite{Zhong2020GeneratingLCP}  &0.144&0.054&0.054&0.328&1.337&0.467&0.445&2.476&5.916&1.079&4.321&9.185\\
        & & RILS\cite{Liu2017PlanningDF}  &0.044&0.017&0.020&0.113&0.346&0.165&0.120&0.658&1.649&0.431&0.920&2.817\\
        & & FIRI(SI) & \bf  0.020& \bf 0.007& \bf 0.008& \bf 0.041& \bf 0.149& \bf 0.058& \bf 0.046& \bf 0.341& \bf 0.544& \bf 0.173& \bf 0.310& \bf 1.049\\
        \hline
        \multirow{6}{*}{\makecell[c]{Line}} & \multirow{3}{*}{\makecell[c]{2-D}} & \textbf{FIRI}  &0.038&0.014&0.010&0.068&0.120&0.039&0.047&0.204&0.263&0.086&0.040&0.440\\
        & & RILS\cite{Liu2017PlanningDF}  &0.014&0.004&0.007&0.028&0.041&0.012&0.016&0.073&0.085&0.028&\bf 0.030&0.142\\
        & & FIRI(SI)  & \bf 0.008& \bf 0.003& \bf 0.002& \bf 0.018& \bf 0.029& \bf 0.010& \bf 0.011& \bf 0.050& \bf 0.079& \bf 0.017& 0.045& \bf 0.115\\
        \cmidrule(lr){2-15} 
        & \multirow{3}{*}{\makecell[c]{3-D}} & \textbf{FIRI}  &0.163&0.037&0.090&0.239&0.678&0.208&0.272&1.110&2.977&0.930&1.709&4.651\\
        & & RILS\cite{Liu2017PlanningDF}  &0.061&0.020&0.027&0.114&0.417&0.134&0.148&0.721&2.069&0.532&1.099&3.214\\
        & & FIRI(SI)  & \bf 0.025& \bf 0.007& \bf 0.012& \bf 0.039& \bf 0.148& \bf 0.049& \bf 0.053& \bf 0.249& \bf 0.696& \bf 0.216& \bf 0.349& \bf 1.084\\
        \hline
        \toprule
    \end{tabular}
\end{table*} 

\subsubsection{Efficiency and Quality}
We conduct experiments to compare efficiency and quality in the aforementioned random environments.
using the same settings of obstacles and boundary as Sec.~\ref{sec::manageability result}. 
\rv{We record the computation time for each algorithm with point seed.}
If IRIS generates a convex polytope during the iteration process that does not include the seed, we force IRIS to terminate prematurely and return the polytope from the previous iteration that includes the seed as the result.
Since both RILS and FIRI guarantee manageability for line seeds, we document their respective time overhead when the seed input is a line as well.
\rv{In addition, since the non-iterative method RILS essentially represents a single iteration of IRIS without MVIE computation, we also record the outcomes of a single iteration of FIRI, denoted as \textbf{FIRI(SI)}.}

\begin{figure}[t]
	\centering
	\includegraphics[width=1\linewidth]{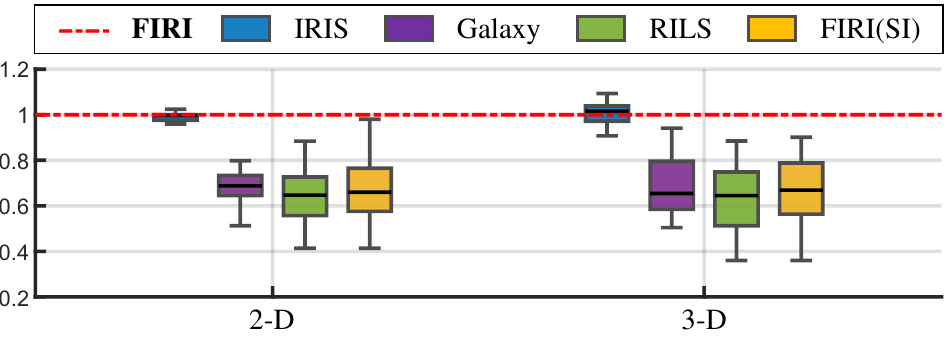}
	\vspace{-0.6cm}
	\caption{
		\rv{Comparison of the sizes of the free convex polytopes generated by different methods.
        The dashed line with a value of 1 indicates the result obtained by FIRI, which we  take as the baseline.}
	}
	\label{fig:result convex hull vol}
\end{figure}

To provide an intuitive comparison of the size of convex polytopes generated by each algorithm, we take the proposed FIRI, which aims to maximize the convex hull volume, as the baseline. 
Specifically, we report the ratio of the volume of the convex hull obtained by each algorithm to the volume of the one obtained by FIRI, as demonstrated in Fig.~\ref{fig:result convex hull vol}.
The efficiency results are presented in Tab.~\ref{tab:Computation Time of convex hull}, where, for clarity, we distinguish between iterative and non-iterative methods using dashed lines in the case of point seed.
In addition, since the number of input obstacles is the primary factor influencing the computational efficiency, we showcase the quantity of input obstacles in different density environments in Tab.~\ref{tab::input obstacle} to provide readers with an insight into the computational efficiency of each algorithm.
\rv{As the results illustrated in Fig.~6}, both IRIS and FIRI iteratively compute larger convex polytopes by continuously inflating the MVIE, resulting in similar sizes. 
However, in IRIS, the SDP-based MVIE solving method consumes significant computation time~\cite{Deits2015ComputingIRIS}, whereas FIRI achieves significant efficiency improvements in MVIE calculations as shown in Sec.~\ref{sec:Comparison of Solving MVIE}, leading to remarkably higher efficiency compared to IRIS \rv{as shown in Tab.~\ref{tab:Computation Time of convex hull}}.
Moreover, FIRI achieves a computational time that is within three times the time of non-iterative RILS without involving MVIE.
The other three non-iterative methods yield smaller polytopes, but achieve better efficiency than IRIS in Tab.~\ref{tab:Computation Time of convex hull}.  
Galaxy, however, is even less efficient than FIRI due to the fact that Quickhull~\cite{barber1996quickhull} involved degrades in the scene to which Galaxy corresponds, which is particularly noticeable in the 3D case.
As for FIRI(SI), similar to RILS, it directly updates the convex hull and can generate polytope of comparable size to RILS, but with higher computational efficiency. 
Moreover, benefiting from the manageability brought about by the RsI, if the objective is to obtain a free convex polytope that includes the seed as fast as possible, without pursuing maximum volume, we believe that FIRI(SI) is a more suitable choice.

\begin{figure}[t]
	\centering
	\includegraphics[width=1\linewidth]{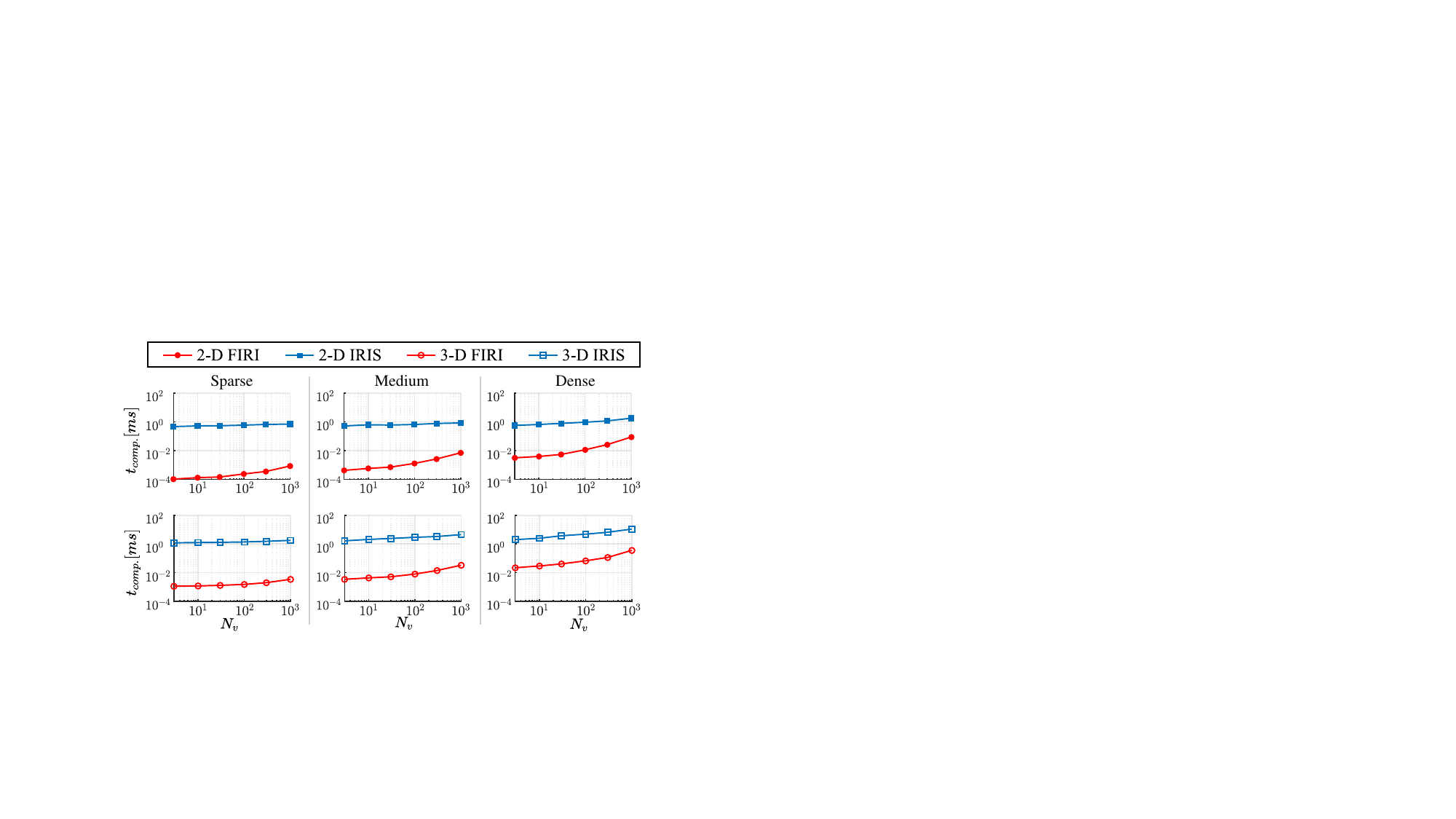}
    \vspace{-0.6cm}
	\caption{
		The benchmark on computation time with the number of obstacle vertices for FIRI and IRIS in different environmental densities.
        The top represents  the 2-D case and the bottom represents the 3-D case.
        FIRI has orders of magnitude advantage in computational efficiency over IRIS.
	}
	\label{fig:result convex hull convex hull}
\end{figure}

Apart from using point representation for obstacles as used above, employing convex polytopes is also a commonly used method for representing the environment. 
\rv{Although methods such as RILS can handle polytope obstacles through discretizing them into points, this approach may incur increased computational burden when discretizing large-scale simple convex polytope, and certain discretization methods (like gridding) may compromise environmental modeling accuracy. 
Hence, we consider the capability to directly process polytope obstacles to be valuable.}
Among above methods in Tab.~\ref{tab:input capablilty}, only FIRI and IRIS support obstacle inputs in the form of convex hulls, thus we further compare these two algorithms.
For each test we generate a certain number of randomly distributed obstacles in a closed space and select the center of the space as the seed input. 
This closed space serves as the boundary.
We conduct benchmarks by setting the number $N_v$ of the vertices of one obstacle to vary from $3$ to $10^3$. 
Compared to the case using point obstacles, the relative volumes of obstacles that are polytopes are larger, thus we vary the number of obstacles to vary from $10$ to $10^3$, which we denote as spare, medium and dense in the results shown in Fig.~\ref{fig:result convex hull convex hull}.
For each the number of the vertices of each different number of obstacles, we perform $500$ tests.
The increasing number of obstacles, the number of halfspaces generated for convex hull construction tends to stabilize, thus the computational efficiency advantage provided by the porposed MVIE solution gradually diminishes.
However, benefiting from the efficient SDMN we propose for halfspace computation, 
FIRI can still maintain a significant advantage over IRIS, even in scenarios with dense obstacles, with a reduction of over 95\% in computational requirements.

\begin{figure}[t]
    \begin{center}
        \subfigure[\label{fig:traj compare 1} The corridors generated by FIRI and RILS in the random forest and the time optimal trajectories constrained in each corridor.]
        {\includegraphics[width=1.0\columnwidth]{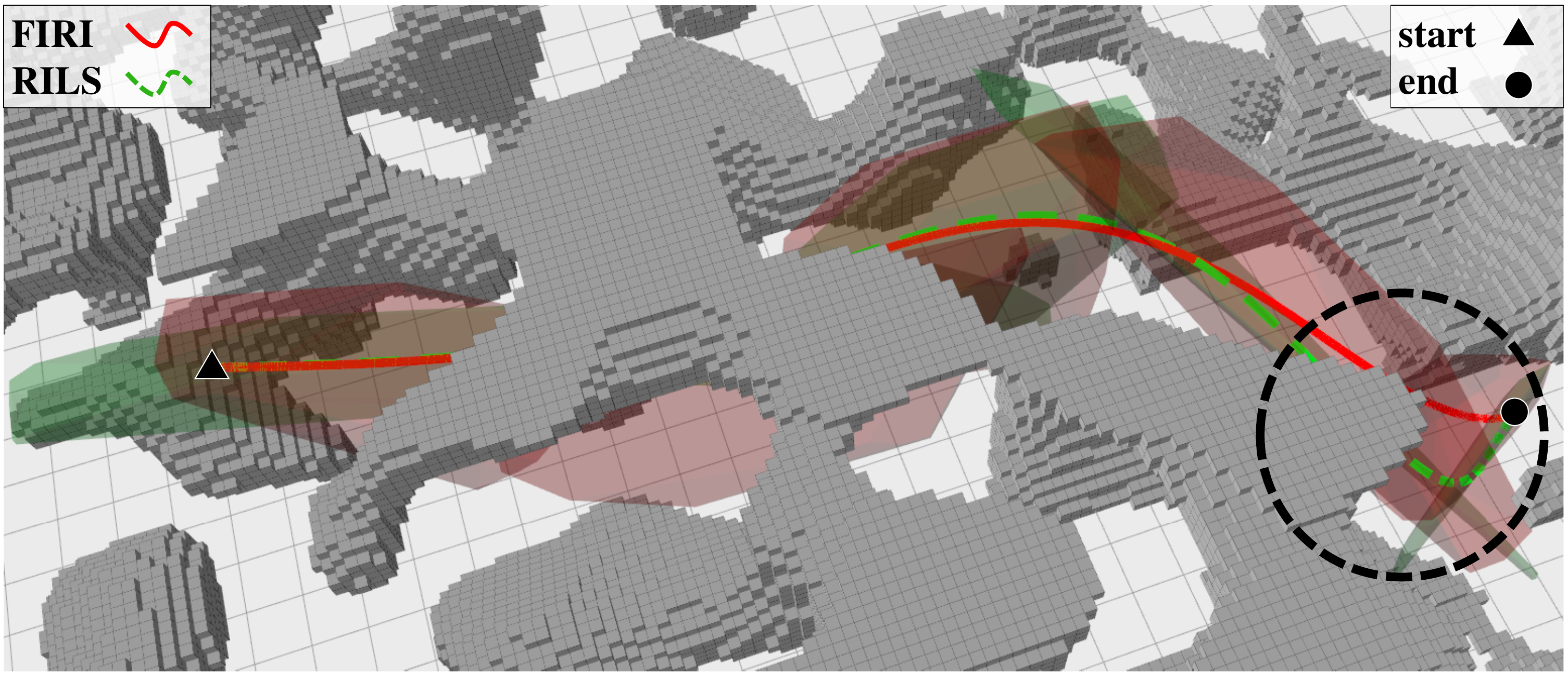}}
        \subfigure[\label{fig:traj compare 2} The speed profiles of the trajectories, colored by the speed magnitude. FIRI generates larger convex polytopes providing more spatial freedom for trajectory optimization.]
        {\includegraphics[width=1.0\columnwidth]{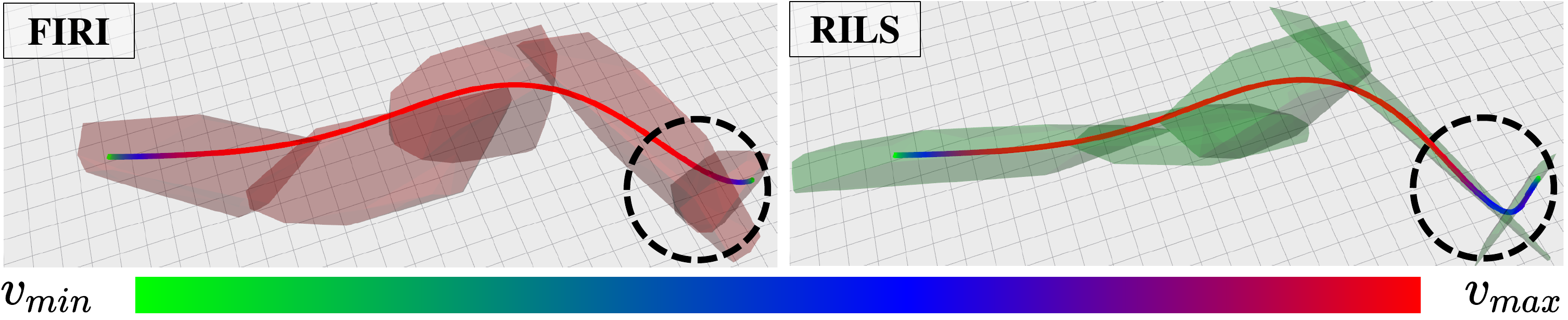}}
        \subfigure[\label{fig:traj compare 3} The velocity and acceleration magnitude for the trajectories constrained in the corridors generated by FIRI and RILS respectively. Greater spatial freedom provided by FIRI leads to more aggressive trajectory performance.]
        {\includegraphics[width=1.0\columnwidth]{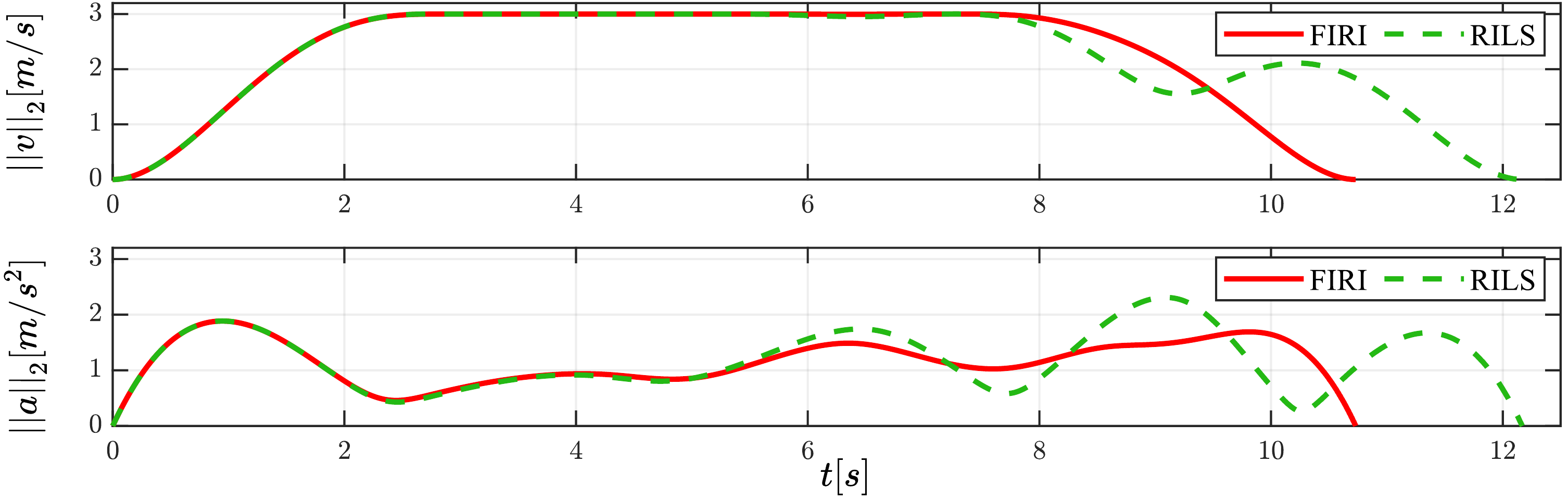}}
    \end{center}
    \vspace{-0.4cm}
    \caption{\label{fig:traj compare} The comparison between the corridor established based on FIRI and RILS in a complex environment and the optimal trajectories constrained within the generated corridor.}
\end{figure}

\subsubsection{Case Study of Quality}
\label{sec:Case Study of Quality}
To demonstrate the significance of convex hull quality and its impact on trajectory planning, we conduct an experiment on trajectory planning based on different convex hull generation methods in the random environment as shown in Fig.~\ref{fig:traj compare 1}.
\rv{Using RRT*\cite{karaman2011sampling}, we generate a collision-free path between start and end points.}
This path can be viewed as a set of connected line segments, based on which we generate convex hulls subsequently.
As indicated in Tab.~\ref{tab:success rate of manageability}, only FIRI and RILS exhibit manageability over line seed.
Thus we build corridors for comparison based on these two methods, which correspond to iterative and non-iterative strategies, regarding the pursuit of maximum volume, respectively. 
The process of generating a safe flight corridor is as follows: 
We sequentially traverse each line segment on the path. 
If the line segment is already included in the previously generated polytopes, we move on to the next segment. 
Otherwise, we use this line as a seed to generate a new convex polytope. 
Due to manageability, the generated set of convex polytopes must have an intersection between two neighboring pairs, forming a corridor.
Based on the generated corridor, we adopt GPOPS-II~\cite{patterson2014gpops} to obtain the optimal trajectory constrained within the corridor.
This collocation-based method transcribes the trajectory optimization into a constrained Nonlinear Programming using the Gauss pseudospectral method, which is then solved by the well-established NLP solver SNOPT~\cite{gill2005snopt}.
In GPOPS-II, each trajectory phase is confined within one polytope, and we set the feasibility constraints for velocity and acceleration as $3m/s$ and $6m/s^2$, and time weight as $20$.
As depicted in Fig.~\ref{fig:traj compare 2}, RILS generates narrow polytopes in the area marked by black dashed circles, resulting in limited maneuvering space for trajectory optimization. 
Consequently, the trajectory constrained in the corridor generated by RILS exhibits a conservative behavior in the marked area . 
In contrast, FIRI, due to its pursuit of maximizing convex hulls, is capable of generating larger corridors, providing greater spatial flexibility in trajectory planning.

\begin{figure}[t]
	\centering
	\includegraphics[width=1\linewidth]{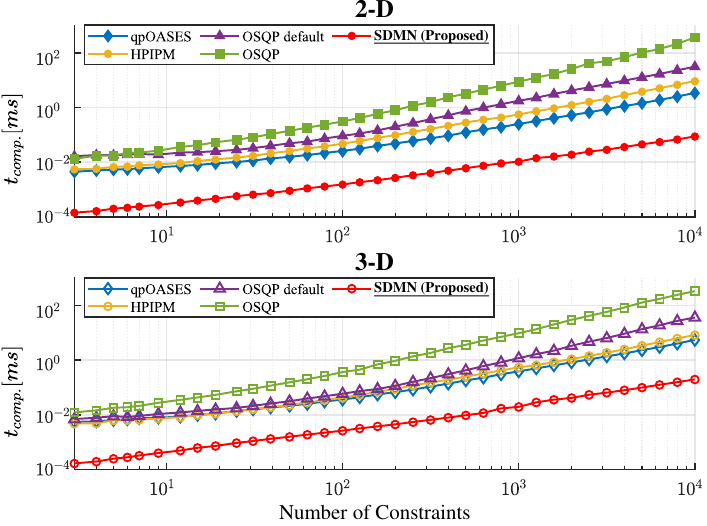}
	\caption{
		Computation time $t_{comp.}$ of different methods for solving strictly convex small dimensional QP under different constraint numbers.
        The proposed SDMN outperforms other methods by orders of magnitudes.
	}
	\label{fig:result QP}
\end{figure}

\begin{table}[t]
    \renewcommand\arraystretch{1.2}
    \tabcolsep=0.07cm
    \caption{Comparison of Precisions $\psi_N$ between Different Methods for Solving Small Dimensional Minimum-Norm}
    \label{tab:cv of QP}
    \centering
    \begin{tabular}{c c c c c c}
        \toprule
        \multirow{3}{*}{{Scenario}} & \multicolumn{5}{c}{Precisions $\psi_N$}  \\
		\cmidrule(lr){2-6} 
        & qpOASES\cite{ferreau2014qpoases} & \makecell[c]{OSQP\cite{osqp}\\default} & OSQP\cite{osqp} & HPIPM\cite{frison2020hpipm} & \makecell[c]{\textbf{SDMN}\\\textbf{(Proposed)}} \\
        \toprule
        2-D   & 1.28e-15 & 5.05e-2 & 6.61e-11 & 7.79e-06 & \textbf{2.78e-17} \\
        3-D   & 2.91e-15 & 2.43e-2 & 2.95e-11 & 4.41e-06 & \textbf{3.55e-17} \\
        \toprule
    \end{tabular}
\end{table}

\subsection{Comparison of Solving Strictly Convex Small Dimensional Minimum-Norm with Massive Constraints}
For minimum-norm (\ref{eq:exampleL2}), we compare SDMN with several cutting-edge general-purpose QP solvers:
the parametric active-set algorithm qpOASES~\cite{ferreau2014qpoases}, 
the operator-splitting based first-order approximation algorithm OSQP~\cite{osqp} and 
the interior-point method-based quadratic programming approximation algorithm HPIPM~\cite{frison2020hpipm}.
We use the open source code~\footnote{\url{https://github.com/coin-or/qpOASES}}\footnote{\url{https://github.com/osqp/osqp}}\footnote{\url{https://github.com/giaf/hpipm}} of these solvers to obtain the solution of (\ref{eq:exampleL2}) respectively.
For qpOASES and HPIPM, we use the default parameters. However, we find that the default parameters of OSQP are insufficient when the constraint size becomes large.
Therefore, we adjust the relative convergence tolerance to $10^{-12}$ and the maximal number of iterations to $10^6$, denoted as OSQP, while the implementation with the default parameters are denoted as OSQP default.
As for the proposed SDMN, we do not require any additional parameter settings.

We compare the solving time of each method, as well as their average precision which is defined as
\begin{equation}
    \label{eq:precision}
    \psi_N=\mathcal{L}_\psi \rbrac{Ey^*- f},
\end{equation}
where $y^*$ is the corresponding solution of each method, $E$ and $f$ are the parameters used to define the constraints of the minimum-norm (\ref{eq:exampleL2}), and $\mathcal{L}_\psi(\cdot)$ denotes a function that takes the absolute value of the largest element of the input vector.
The precision $\psi_N$ indicates the degree of proximity of the solution of each method to the most active constraints of the input minimum-norm.
The performance is reported in Fig.~\ref{fig:result QP} and Tab.~\ref{tab:cv of QP}.
The results demonstrate that SDMN exhibits remarkably high computational efficiency, surpassing other methods by orders of magnitude. 
Additionally, SDMN provides analytical solutions, achieving high accuracy without the need for additional parameters.

\subsection{Comparison of Solving MVIE}
\label{sec:Comparison of Solving MVIE}

For MVIE (\ref{eq:AbstractMVIE}), we propose two algorithms, namely the SOCP-reformulation algorithm and the randomized algorithm specialized for 2-D case. 
The randomized algorithm yields an analytical solution, here we abbreviate it as \textbf{RAN}.
To distinguish from above methods, we denote the proposed SOCP-reformulation algorithm which is solved by affine scaling method as \textbf{CAS}.
We compare them with three methods: 
1) The optimization method based on SDP formulation of MVIE in IRIS~\cite{Deits2015ComputingIRIS}. 
2) The example~\footnote{\url{https://docs.mosek.com/latest/cxxfusion/examples-list.html\#doc-example-file-lownerjohn-ellipsoid-cc}} of the cutting-edge solver Mosek~\cite{MosekOPT} to computes the Lowner-John inner ellipsoidal approximations of a polytope. 
3) A strategy of solving the SOCP form~(\ref{eq:SOCPMVIE}) of the MVIE by Mosek. 
As the example in Mosek formulate MVIE into a mixed conic quadratic and semidefinite problem, we denote it as \textbf{Mosek~SDP} and denote another strategy using Mosek as \textbf{Mosek~SOCP}.
We use the default parameters for IRIS and Mosek.

\begin{figure}[t]
	\centering
	\includegraphics[width=1\linewidth]{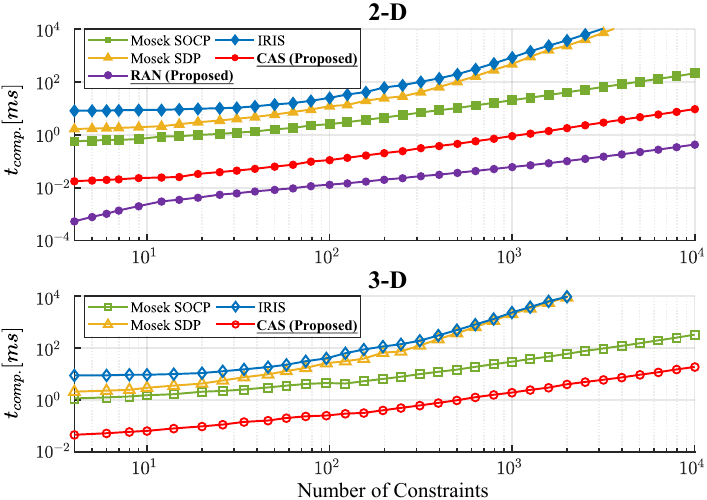}
	\caption{
		\rv{Computation time $t_{comp.}$ of different methods for solving MVIE under different constraint numbers.
        The proposed CAS outperforms other methods by orders of magnitudes.
        Additionally, the proposed analytical method RAN further enhances efficiency for the 2D case.}
	}
	\label{fig:result MVIE}
\end{figure}

We compare the computation time and average precision of each method to calculate the maximum ellipsoid in closed convex polytopes consisting of different numbers of halfspaces.
The precision is defined as 
\begin{equation}
    \psi_\mathcal{E}=\mathcal{L}_\psi \left(\sbrac{\sbrac{A_\mathcal{P}{A_\mathcal{E}}^*{D_\mathcal{E}}^*}^2\mathbf{1}}^{\frac{1}{2}} +A_\mathcal{P}{b_\mathcal{E}}^* - b_\mathcal{P} \right),
\end{equation}
where ${A_\mathcal{E}}^*$, ${D_\mathcal{E}}^*$ and ${b_\mathcal{E}}^*$ are the coefficients of the maximum inscribed ellipsoid solved by each method, $A_{\mathcal{P}}$ and $b_{\mathcal{P}}$ are the parameters that define the halfspaces of the input polytope $\mathcal{P} $ defined in (\ref{eq:convex_polytope}), and $\mathcal{L}_\psi(\cdot)$ is the function defined in (\ref{eq:precision}).
This precision $\psi_\mathcal{E}$ represents the proximity of the ellipsoids obtained by different methods to the most active halfspace constraint, which are constructed in the form of second-order cone constraint as~(\ref{eq::OrthonormalityMVIE_constraints}).
The results are summarized in Fig.~\ref{fig:result MVIE} and Tab.~\ref{tab:cv of MVIE}.
By comparing the performance of Mosek SDP and Mosek SOCP, we can get the conclusion that transforming the commonly used SDP formulation to the SOCP formulation, which is presented in Sec.~\ref{sec:SOCP-Reformulation of MVIE}, leads to a significant improvement in computational efficiency without sacrificing precision.
Additionally, in the low-dimension massive-constraint case faced in this paper, the use of the affine scaling method avoids the requirement of solving a large-scale system of linear equations at each iteration, compared to the primal-dual interior point method used in Mosek.
Therefore as the results demonstrate, CAS is capable of solving MVIE in less time while maintaining a comparable precision. 
Moreover, for the 2-D case, our proposed linear-time complexity algorithm RAN further enhances the computational efficiency by orders of magnitude and compute analytical results.

\begin{table}[t]
    \renewcommand\arraystretch{1.2}
    \tabcolsep=0.13cm
    \caption{Comparison of Precision $\psi_\mathcal{E}$ between Different Methods for Solving MVIE}
    \label{tab:cv of MVIE}
    \centering
    \begin{tabular}{c c c c c c}
        \toprule
        \multirow{3}{*}{\centering {Scenario}} & \multicolumn{5}{c}{Precisions $\psi_\mathcal{E}$}  \\
		\cmidrule(lr){2-6} 
        & IRIS\cite{Deits2015ComputingIRIS} & \makecell[c]{Mosek\cite{MosekOPT}\\SDP} & \makecell[c]{Mosek\cite{MosekOPT}\\SOCP} & \makecell[c]{\textbf{CAS}\\\textbf{(Proposed)}} & \makecell[c]{\textbf{RAN}\\\textbf{(Proposed)}}  \\
        \toprule
        2-D & 8.56e-8 & 6.47e-9 & 4.87e-12 & 1.59e-8 & \textbf{4.41e-16} \\
        3-D & 1.54e-8 & 1.56e-8 & 4.05e-12 & 2.04e-8 & / \\
        \toprule
    \end{tabular}
\end{table}

\section{Real-world Application}

\begin{figure}[t]
	\centering
	\includegraphics[width=1\linewidth]{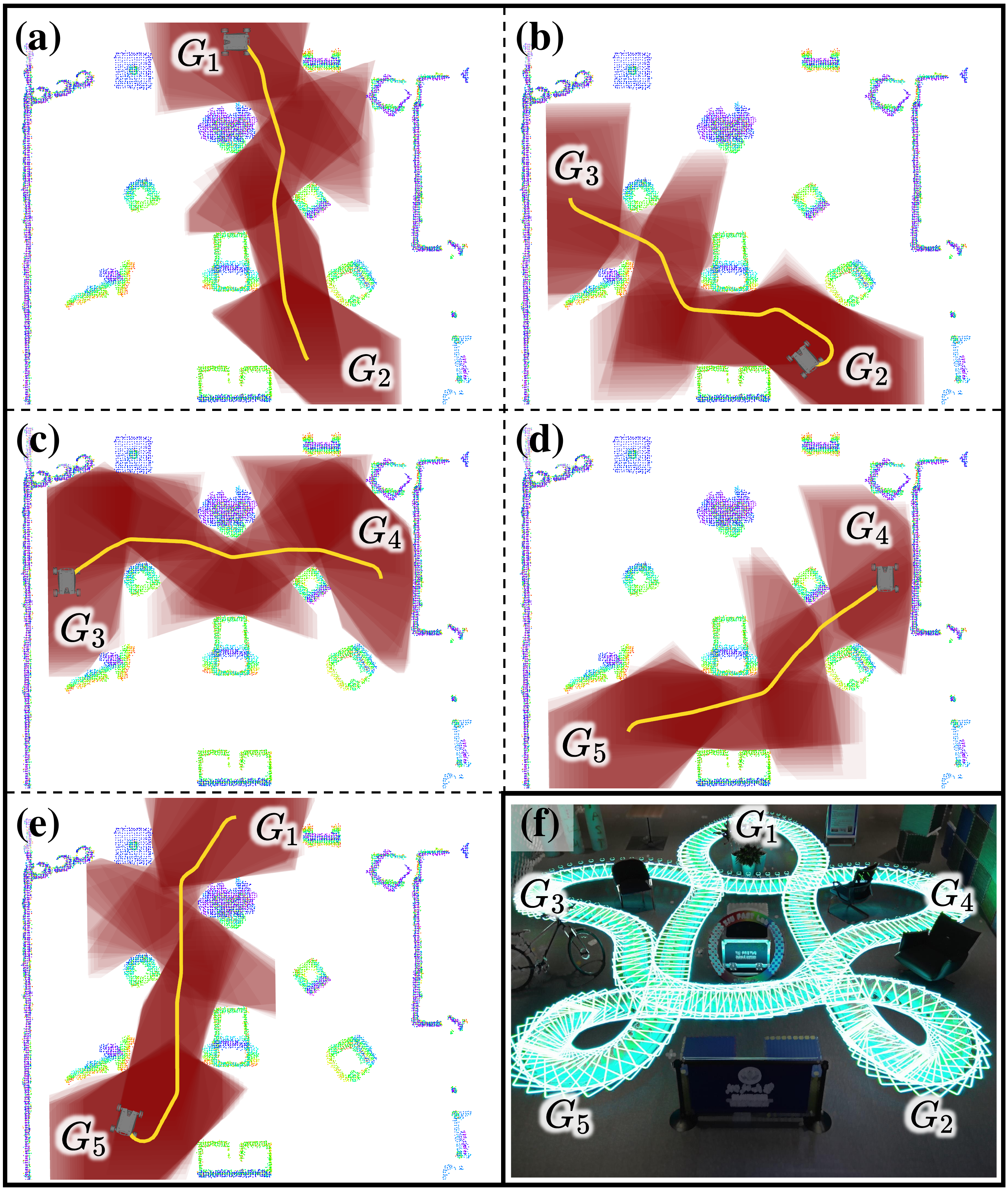}
	\vspace{-0.6cm}
	\caption{
		\textbf{(a)-(e)}: Illustration of the corridors generated by FIRI during the traversal of the robot from $G_1$ through $G_2$-$G_5$ and finally back to $G_1$ in a cluttered environment. 
        \textbf{(f)}: Snapshots of the robot's navigation result.
	}
	\label{fig:pentagram}
\end{figure}

\begin{figure*}[t]
    \begin{center}
        \subfigure[\label{fig:maze1} \textbf{Left}: Illustration of the generation of a dense corridor based on the yellow rough trajectory obtained from the front-end. \textbf{Right}: We optimize a smooth safe whole-body trajectory by constraining the trajectory in the generated corridor in the back-end.]
        {\includegraphics[width=1.95\columnwidth]{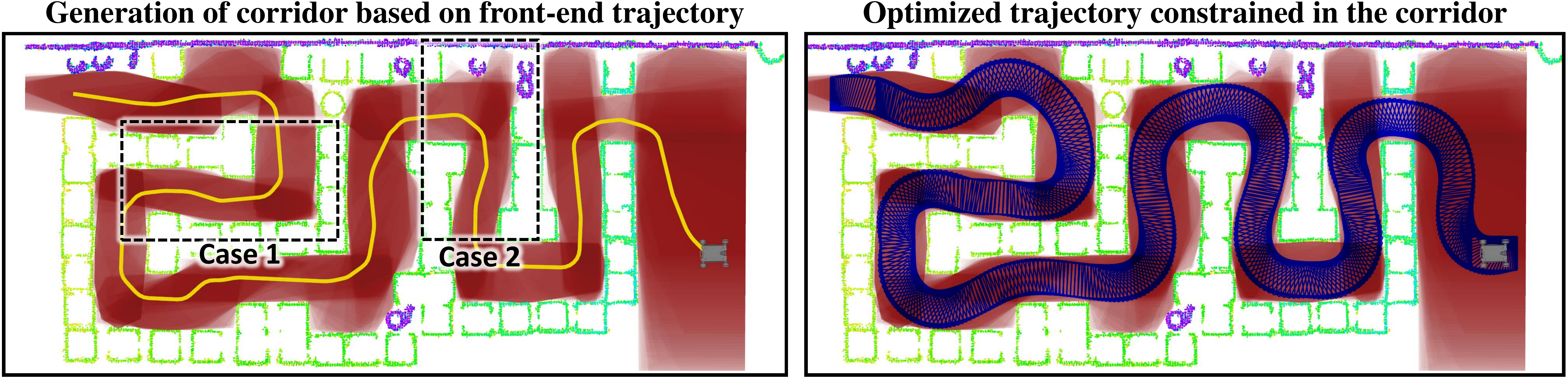}}
        \vspace{-0.4cm}
        \subfigure[\label{fig:maze2} \textbf{Left}: 
        For the two specific cases in (a), we demonstrate the convex polytopes generated by the different methods. The results show that only FIRI is able to satisfy the whole-body planning requirement of generating polytopes that fully contain the robot. \textbf{Right}: Snapshots of the robot's navigation result.]
        {\includegraphics[width=1.95\columnwidth]{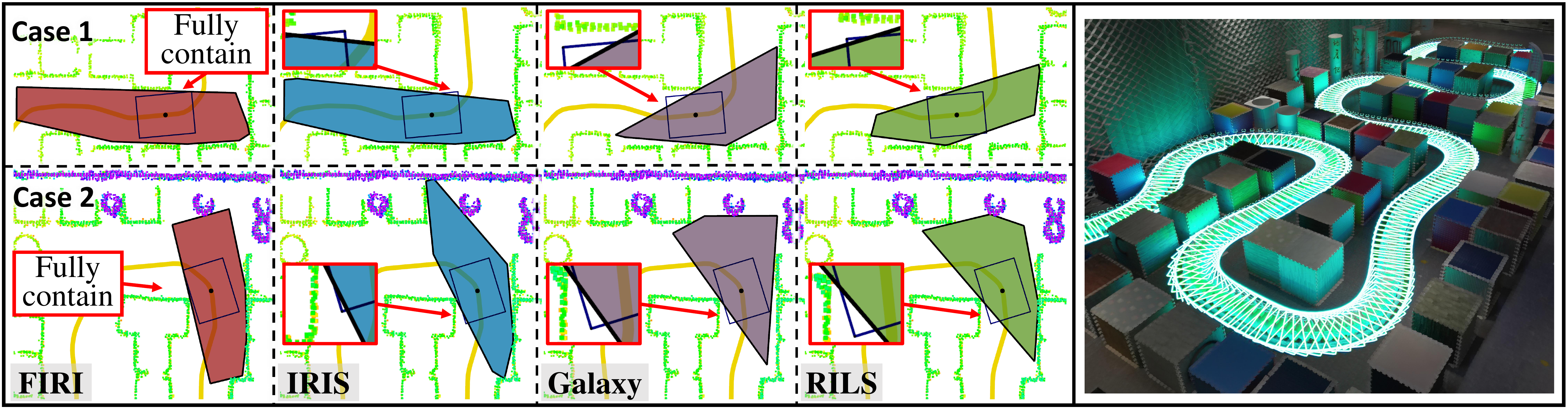}}
    \end{center}
    \caption{\label{fig:maze} A real-world application of a differential driven robot tasked with traversing a maze, utilizing FIRI to abstract the feasible region.}
    \vspace{-0.4cm}
\end{figure*}

To validate the performance of FIRI in practical applications, we designed two real-world applications: a global trajectory planning for a vehicle robot with nonholonomic constraints in 2-D environment and a local planning for a quadrotor in 3-D environment.

\subsection{Dense Corridor for 2-D Whole-body Planning}
The trajectory planning for large-volume vehicle in narrow environments requests the entire vehicle to be constrained within the safe region, which puts a demand on manageability.

We adopt a differential driven AgileX SCOUT MINI~\footnote{\url{https://global.agilex.ai/products/scout-mini}} as the physical platform. 
The platform is equipped with an Intel NUC~\footnote{\url{https://www.intel.com/content/dam/support/us/en/documents/intel-nuc/NUC11PH_TechProdSpec.pdf}} with an Intel Core i7-1165G7 as onboard processor and a LiDAR sensor~\footnote{\url{https://ouster.com/products/hardware/os1-lidar-sensor}} integrated with an IMU for perception. 
We use FAST-LIO2~\cite{xu2022fast} for both pre-mapping the environment and real-time localization during execution of the planned trajectory. 
That is, the obstacles are represented by points.

Building on our previous work~\cite{han2023efficient}, we use hybrid A* as the front-end to obtain an initial feasible trajectory, and then finely sample on the trajectory. 
The robot shape corresponding to each sampled state is used as a seed to generate a convex polytope by FIRI, forming a dense safety corridor, as shown in Fig.~\ref{fig:maze}. 
In the back-end, we utilize polynomial representation for the trajectory and constrain that the vehicle shape corresponding to each sampled constraint state on the trajectory is fully contained within the corresponding convex polytope of the safety corridor. 
Considering localization errors and control errors, for safety purposes, we inflate the robot's shape during planning to generate a safer corridor and trajectory.

\subsubsection{Cluttered Environment}
\rv{The results are shown in Fig~\ref{fig:pentagram}. 
We require the robot to navigate through a cluttered environment filled with irregular obstacles, starting from $G_1$ and sequentially reaching $G_2$-$G_5$, and finally returning to $G_1$. 
Since the focus of this paper is on the generation of free convex polytope, only the front-end trajectory that guides the generation of convex polytope and the generated convex polytopes are demonstrated. 
We refer the readers to the accompanying video for the detailed navigation process. In this experiment, as shown in Fig~\ref{fig:pentagram}, we generate $5$ dense corridors, with an average of $91.8$ convex polytopes in each corridor. 
For each convex polytope, the average number of input obstacle points was $715.3$, with an average processing time of $0.131~ms$. 
The average time taken to generate a corridor is $12.02~ms$.}

\subsubsection{Narrow Maze}
\label{sec:application car maze}
\rv{We task the robot with traversing through a narrow maze. 
The result, as depicted in Fig.~\ref{fig:maze1}, demonstrates that FIRI, based on the yellow front-end trajectory, generates high-quality convex polytopes, providing sufficient spatial degrees of freedom for trajectory optimization in the back-end. 
Additionally, as shown in Fig.~\ref{fig:maze2}, to intuitively demonstrate the manageability of our experiment, we select two particularly narrow cases within the maze environment, illustrating the results of the convex polytopes generated by various methods~\cite{Deits2015ComputingIRIS,Zhong2020GeneratingLCP,Liu2017PlanningDF} compared in Sec.~\ref{sec:Comparison of Generating Free Convex Polytope}. 
It can be observed that only FIRI is capable of fully enclosing the robot. 
A more detailed demonstration of the generation from each seed and its corresponding convex polytope along the entire trajectory is presented in the accompanying video. 
In this experiment, we generate a dense corridor consisting of $550$ convex polytopes. 
For each convex polytope, the average number of input obstacle points is $1307.1$, with an average processing time of $0.249~ms$. 
The total time taken for generating the entire corridor is $136.74~ms$.}

\begin{figure}[t]
    \begin{center}
        {\includegraphics[width=1.0\columnwidth]{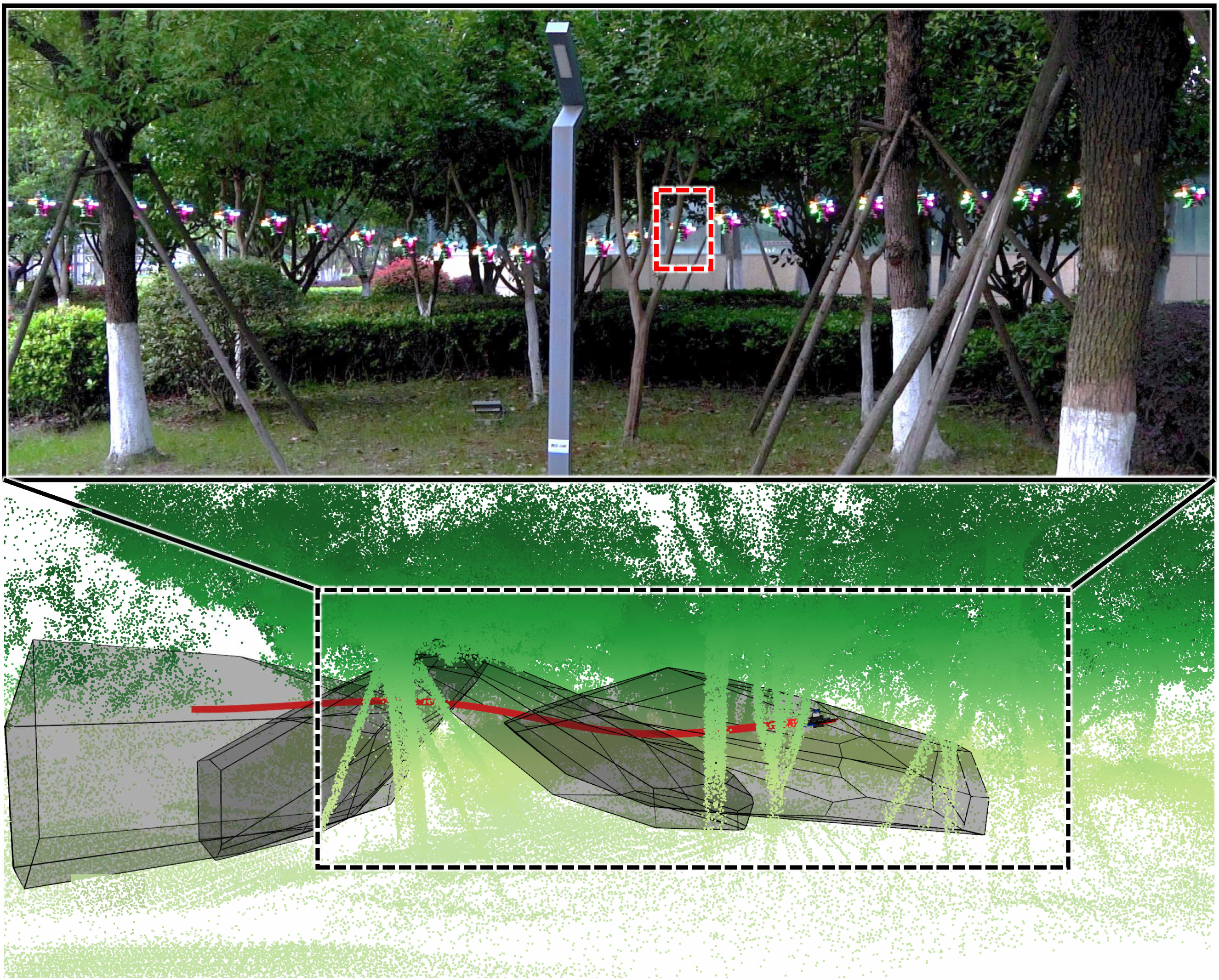}}
    \end{center}
    \vspace{-0.4cm}
    \caption{\label{fig:toutu toutu} Illustration of the application of FIRI in a local trajectory planning framework for a quadrotor to navigate through a cluttered forest.
    \textbf{Top}: A snapshots of the quadrotor traveling through the forest.
    \textbf{Bottom}: The visualization of quadrotor replanning for the moments corresponding to the red boxes.
    The black transparent polytopes are generated by FIRI.}
    \vspace{0.2cm}
\end{figure}

\subsection{Sparse Corridor for 3-D Local Replanning}

Our platform is a customized quadrotor equipped with an NVIDIA Orin NX\footnote{\url{https://www.nvidia.com/en-us/autonomous-machines/}} as the onboard processor and a Livox MID-360\footnote{\url{https://www.livoxtech.com/mid-360}} LiDAR for perception.
For the quadrotor, due to its small size, we model it as a point mass, similar to many other works~\cite{Liu2017PlanningDF,Gao2020TeachRepeatReplanAC}.
We employ  FAST-LIO2~\cite{xu2022fast} for online state estimation and utilize occupancy grid map to filter sensor noise, which is easy to inflate for safety margins.
For trajectory planning, the path generation and corridor generation are the same as Sec.~\ref{sec:Case Study of Quality}. Then building upon our previous work~\cite{wang2022geometrically}, we utilize piecewise polynomials to represent the trajectory. Each piece of the trajectory is constrained within the corresponding polytope to ensure safety.

As shown in Fig.~\ref{fig:toutu toutu}, the experiment is conducted in a dense forest, and the quadrotor is required to navigate through a series of waypoints.
Since the environment is unknown, we make the quadrotor perform high-frequency replan (20Hz) according to the real-time perception.
The maximum velocity of the UAV during flight is up to 4.5 m/s. 
Each replan involves 3-7 convex polytopes, the average generation time for each polytope is $2.76~ms$, and the average number of input obstacle points for FIRI is $8219.6$.
We refer readers to the accompanying video for more information.

\section{Discussion and Conclusion}

Our algorithm FIRI aims to generate the largest possible convex polytopes. However, when it comes to practical applications such as trajectory planning, we cannot guarantee that the inflation direction of polytope is always favorable for trajectory optimization. 
For example, when the robot has a high velocity, if the inflation direction is perpendicular to the velocity direction, it may lead to generated polytope that are not conducive to trajectory optimization. 
Our future research direction will focus on generating convex polytopes that are  application-friendly.
\rv{Additionally, the number of faces of the generated polytope significantly affects how fast the motion planning problem can be solved. 
As discussed in Sec.~III.B1, we currently employ a widely used greedy method to reduce this number,
and in the future, we will explicitly incorporate the face number as a further criterion of the region quality, while balancing it with the volume of the polytope. }

In conclusion, we propose a novel obstacle-free convex polytope generation algorithm called FIRI, which achieves high quality, efficiency, and manageability simultaneously. 
To achieve efficiency, we design targeted methods for the two optimization problems involved in FIRI. 
Specifically, for 2-D MVIE, we develop a linear-time complexity method, which is proposed for the first time. 
We perform extensive benchmarks against several convex polytope generation algorithms to confirm the superior performance of FIRI.
The comparisons with general-purpose solvers demonstrate the speedup over orders of magnitude of our targeted methods.
Two typical applications showcase the practicality of FIRI.

\rv{CGAL~\cite{fabri2009cgal} lacks a dedicated solver for 2-D MVIE.} In the future, we will continue our research on 2-D MVIE and work towards implementing a version of rational predicates.

\section{Acknowledgment}
The linear-time complexity algorithm for 2-D MVIE is initially designed by Zhepei Wang and finalized by Qianhao Wang.
The authors would like to thank Yuan Zhou and Mengze Tian for their help in the hardware of the 2-D application.

\appendix
We provide a proof of monotonicity and locality of the proposed value function~(\ref{eq:value_function}) for 2-D MVIE in Sec.~\ref{sec:Value Function}.

\vspace{-0.3cm}
\subsection{Proof of Monotonicity}
\label{apd:Proof of Monotonicity}
\begin{proof} 
    $\mathcal{G}$ is unclosed: 
    For the polar cone of a set, when we add elements to the set, the angle of the cone either remains the same or decreases. 
    Therefore, the first part of the function $\bar{w}_u(\mathcal{G})$ is monotonically increasing. Additionally, the second part of the function is clearly monotonically increasing. 
    Since we combine these two parts using the lexicographical ordering $<$ , the function $\bar{w}_u(\mathcal{G})$ satisfy monotonicity.

    $\mathcal{G}$ is closed: 
    Since the second element of the function $\bar{w}_c(\mathcal{G})$ is constant, we only need to prove the monotonicity of the first part.
    We denot a constraint $h\notin \mathcal{G}$.
    The MVIE of $\mathcal{G}\cup\{h\}$ certainly satisfies the original constraints of $\mathcal{G}$, thus its area must be less than or equal to the area of the MVIE of $\mathcal{G}$. 

    As stated in Sec.~\ref{sec:Value Function}, the first element of $\bar{w}_c$ is always greater than 0, and the first element of $\bar{w}_u$ is always less than or equal to 0, thus we have $\bar{w}_c > \bar{w}_u$.
\end{proof}

\vspace{-0.4cm}
\subsection{Proof of Locality}
\label{apd:Proof of Locality}
\begin{proof}
    For finite sets $\mathcal{G}$ and $\mathcal{F}$ and a constraint $h$ such that $\mathcal{G} \subseteq \mathcal{F} $, $h\notin \mathcal{F}$.
    When $-\infty<w(\mathcal{G})=w(\mathcal{F})$, these two sets must either both be closed or both be unclosed.

    $\mathcal{G}$ is unclosed: 
    Based on the above condition, the polar cones of the new sets formed by the unit normal vectors of the respective elements of $\mathcal{G}$ and $\mathcal{F}$ are the same.
    Thus the new element $h$ makes any change in this cone or even its corresponding normal vector falling in this cone leads to a closed polytope, which is the same for both $\mathcal{G}$ and $\mathcal{F}$.

    $\mathcal{G}$ is closed: 
    Due to the uniqueness of MVIE, if $\mathcal{G} \subseteq \mathcal{F} $ and 
    $w(\mathcal{G})=w(\mathcal{F})$, then their MVIE corresponds to the same ellipse. Thus if $h$ is violated by $\mathcal{F}$, it will be violated by $\mathcal{G}$.

\end{proof}

\bibliography{references}

\end{document}